\documentclass[lettersize,journal]{IEEEtran}
\usepackage{amsmath,amsfonts}
\usepackage{algorithmicx}
\usepackage{threeparttable}
\usepackage{array}
\usepackage[caption=false,font=normalsize,labelfont=sf,textfont=sf]{subfig}
\usepackage{textcomp}
\usepackage{stfloats}
\usepackage{url}
\usepackage{verbatim}
\usepackage{graphicx}
\usepackage{cite}
\usepackage{caption}
\usepackage{amsmath,amssymb,amsfonts}
\usepackage{calligra}
\usepackage[noend]{algpseudocode}
\usepackage{algorithmicx,algorithm}
\usepackage{amsmath} 
\usepackage{graphicx}
\usepackage{textcomp}
\usepackage{soul}
\usepackage{booktabs}
\usepackage{multirow}
\usepackage{multicol}
\usepackage{hyperref}
\usepackage{makecell}
\usepackage{footnote}
\usepackage{xcolor}

\begin{document}

\title{VSR-Net: Vessel-like Structure Rehabilitation Network with Graph Clustering}
\author{Haili Ye$^\dag$,
	Xiaoqing Zhang$^\dag$$^\star$,
	Yan Hu,
	Huazhu Fu,
	and Jiang Liu$^\star$ 
\IEEEcompsocitemizethanks{\IEEEcompsocthanksitem Haili Ye, Xiaoqing Zhang, Yan Hu, and Jiang Liu are with the Research Institute of Trustworthy Autonomous Systems [Department of Computer Science and Engineering], Southern University of Science and Technology, Shenzhen, China.
\IEEEcompsocthanksitem Huazhu Fu is with the Institute of High-Performance Computing (IHPC), Agency for Science, Technology and Research (A*STAR), Singapore. \\
$^\dag$ Haili Ye and Xiaoqing Zhang contribute equally. \\
$^\star$ Xiaoqing Zhang and Jiang Liu are corresponding authors. (11930927@mail.sustech.edu.cn, liuj@sustech.edu.cn)}}%

%
\markboth{IEEE TRANSACTIONS ON IMAGE PROCESSING, VOL. x, 2023}%
{Shell \MakeLowercase{\textit{et al.}}: A Sample Article Using IEEEtran.cls for IEEE Journals}


\maketitle

\begin{abstract}
The morphologies of vessel-like structures, such as blood vessels and nerve fibres, play significant roles in disease diagnosis, e.g., Parkinson’s disease. Deep network-based refinement segmentation methods have recently achieved promising vessel-like structure segmentation results. There are still two challenges: (1) existing methods have limitations in rehabilitating subsection ruptures in segmented vessel-like structures; (2) they are often overconfident in predicted segmentation results. To tackle these two challenges, this paper attempts to leverage the potential of spatial interconnection relationships among subsection ruptures from the structure rehabilitation perspective. Based on this, we propose a novel Vessel-like Structure Rehabilitation Network (VSR-Net) to rehabilitate subsection ruptures and improve the model calibration based on coarse vessel-like structure segmentation results. VSR-Net first constructs subsection rupture clusters with Curvilinear Clustering Module (CCM). Then, the well-designed Curvilinear Merging Module (CMM) is applied to rehabilitate the subsection ruptures to obtain the refined vessel-like structures. Extensive experiments on five 2D/3D medical image datasets show that VSR-Net significantly outperforms state-of-the-art (SOTA) refinement segmentation methods with lower calibration error. Additionally, we provide quantitative analysis to explain the morphological difference between the rehabilitation results of VSR-Net and ground truth (GT), which is smaller than SOTA methods and GT, demonstrating that our method better rehabilitates vessel-like structures by restoring subsection ruptures. 

\end{abstract}


\section{Introduction}
\IEEEPARstart{T}{he} morphologies of vessel-like structures (such as blood vessels, trachea, lymphatic vessels, and nerve fibres) are closely related to the pathogenesis of the disease \cite{Ahn2021}. Accurate vessel-like structure segmentation is significant for auxiliary diagnosis of related diseases \cite{Ali2022}. This is because morphological parameters extracted based on vessel-like structure segmentation results can provide a potential reference for analyzing etiopathogenesis, significantly affecting the clinical diagnosis results. For example, the density and thickness of retinal blood vessels in the fundus are essential indicators for evaluating retinal edema \cite{Suciu2020} and venous occlusion \cite{Amir2017}.

\begin{figure}[!t]
\centerline{\includegraphics[width=0.85\columnwidth]{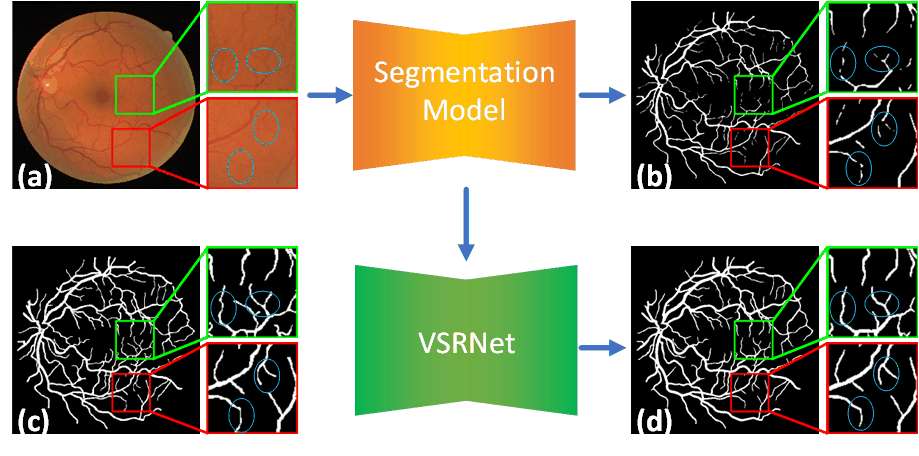}}
	\caption{(a) One representative color fundus image selected from the DRIVE \cite{DRIVE} dataset. (b) coarse vessel-like structure segmentation results of U-Net\cite{UNet}. (c) ground truth. (d) vessel structure rehabilitation results of VSR-Net. From the two enlarged sub-regions on the color fundus image (top left corner), we see subsection ruptures in the coarse vessel-like structure segmentation results of U-Net (marked with a blue circle). VSR-Net effectively rehabilitates these subsection ruptures in vessel-like structures.}
	\label{fig1}
\end{figure}

In the past decades, conventional segmentation methods have widely been utilized to address vessel-like structure segmentation tasks, such as contour model-based method \cite{Chung2018}, threshold-based segmentation methods \cite{LiuWen2022}, and edge detection methods \cite{Ooi2021}. These methods heavily relied on manual feature extraction design and continuous threshold adjustment, which are time-consuming and fragile. Moreover, there are many subsection ruptures in vessel-like structures segmentation results based on conventional segmentation methods. The advent of deep networks and the ongoing progress in medical image segmentation have ushered in a paradigm shift in vessel-like structure segmentation tasks. Deep network-based semantic segmentation methods \cite{UNet, SegNet} have simplified the segmentation process of conventional segmentation methods in an end-to-end manner. Nonetheless, they also bring subsection rupture problems in vessel-like structure segmentation results. Fig.~\ref{fig1}(b) offers vessel-like structure segmentation results generated by U-Net~\cite{UNet} based on color fundus images. We see that there exists a subsection rupture problem in segmented vessel-like structures through comparisons to the two enlarged sub-regions of ground truth (as shown in Fig.~\ref{fig1}(c)). Two possible reasons to explain this phenomenon: (1) Most deep network-based semantic segmentation methods provide category information based on pixel-wise predicted probability values, unable to rectify inconsistent segmentation results around subsection fracture positions \cite{CENet, CS2Net, Sklcon}. (2) Inherent imaging factors like similar density distributions and blurred vessel-like morphologies also bring difficulties for existing methods to segment small vessel structures accurately. In particular, the subsection rupture problem breaks the integrity of vessel-like structures, negatively affecting the disease diagnosis results of clinicians \cite{Ali2022, Samber2020}.

Deep network-based refinement segmentation methods have widely been applied to re-segment/refine coarse segmentation results \cite{DoubleUNet, RFNet, AGLN}. Nonetheless, they only bring marginal improvements in segmentation results at the cost of improving the complexity of deep network architectures. Moreover, they also have limitations in tackling the subsection rupture problem without considering its particularity.
In addition, current deep network-based refinement segmentation methods are over-confident in predicted segmentation results, increasing the risk of clinical diagnosis errors. Expected calibration error (ECE) \cite{guo2017calibration} is a commonly-accepted metric to evaluate the calibration of the network, which has not been adopted in  vessel-like structure segmentation tasks.
Moreover, we find that there exist spatial interconnection relationships among subsection ruptures by observing vessel-like structure segmentation results with U-Net (as shown in Fig.~\ref{fig1}(b). We argue that subordinate relationships of subsection ruptures can be inferred via the end of curvilinear direction. Graph Convolution Networks (GCNs) \cite{GCNSu} have shown powerful feature representation learning ability in possessing graph data, facilitating feature representation propagation between nodes and their neighboring nodes through layered diffusion. Previous works~\cite{GCNSurvey} have demonstrated the effectiveness of GCNs in a variety of tasks, including image segmentation and classification, which have not been exploited to rehabilitate the subsection ruptures in segmented vessel-like structures.

Motivated by the above systematical analysis, we consider the subsection rupture problem as vessel-like structure rehabilitation rather than vessel-like structure segmentation. Thus, we propose a novel Vessel-like Structure Rehabilitation Network (VSR-Net) to rehabilitate subsection ruptures in coarse vessel-like structure segmentation results and improve the confidence calibration, as shown in Fig.~\ref{fig1}(d). In particular, in the preparation stage, we obtain coarse vessel-like structure segmentation results with existing vessel-like structure segmentation baselines, e.g., U-Net. In the rehabilitation stage, we first present a Curvilinear Clustering Module (CCM) to convert subsection ruptures into graphs and construct the clustering relationships among individual subsection ruptures via graph convolution operation, which exploits the potential of spatial interconnection relationships among subsection ruptures. Then, this paper introduces a Curvilinear Merging Module (CMM) to rehabilitate subsection ruptures in coarse segmented vessel-like structures, guided by clustering results generated by the CCM. Moreover, no previous work has systematically evaluated the performance of vessel-like structure rehabilitation from the structure morphology perspective; we are the first to adopt Vascular Bifurcation Number ($VBN$) \cite{VBN}, Fractal Dimension ($FD$) \cite{FD}, and Vascular Tortuosity ($VT$) \cite{VT} to evaluate the vessel-like structure morphology rehabilitation. Compared with dice and Jaccard score, these three structure morphology metrics can objectively measure the morphological differences between the prediction results and ground truth, conforming to clinical diagnosis requirements. For example, subsection ruptures at the bifurcation site reduce the number of vascular bifurcation \cite{VBN}, which may increase the ratio of misdiagnosing abnormal vascular congestion.

We summarize the main contributions of this paper as follows:
\begin{itemize}
\item We are the first to consider the subsection rupture problem as a vessel-like structure rehabilitation task rather than a vessel-like structure segmentation task, providing novel insight into addressing this problem.

\item We propose a novel Vessel-like Structure Rehabilitation Network (VSR-Net) to rehabilitate subsection ruptures and improve confidence calibration based on coarse vessel-like structure segmentation results, fully leveraging the potential of spatial interconnection relationships among subsection ruptures via graph clustering.

\item We systematically evaluated the performance of vessel-like structure rehabilitation from the structure morphology perspective. The experimental results on 2D/3D medical image datasets show that our VSR-Net performs better in terms of vessel-like structure segmentation results, vessel-like structure rehabilitation results, and confidence calibration through comparisons to SOTA methods.

\item We conduct quantitative analysis and ablation study to validate the generalization ability and robustness of our method.

\end{itemize}

This paper is organized as follows: we describe related works in Section 2. Then, we introduce our VSR-Net in Section 3. Next, we conduct ablation studies and comparable experiments in Section 4. Section 5 concludes this paper.

\section{Related Work}

\begin{figure*}[!th]
	\centerline{\includegraphics[width=0.65\linewidth]{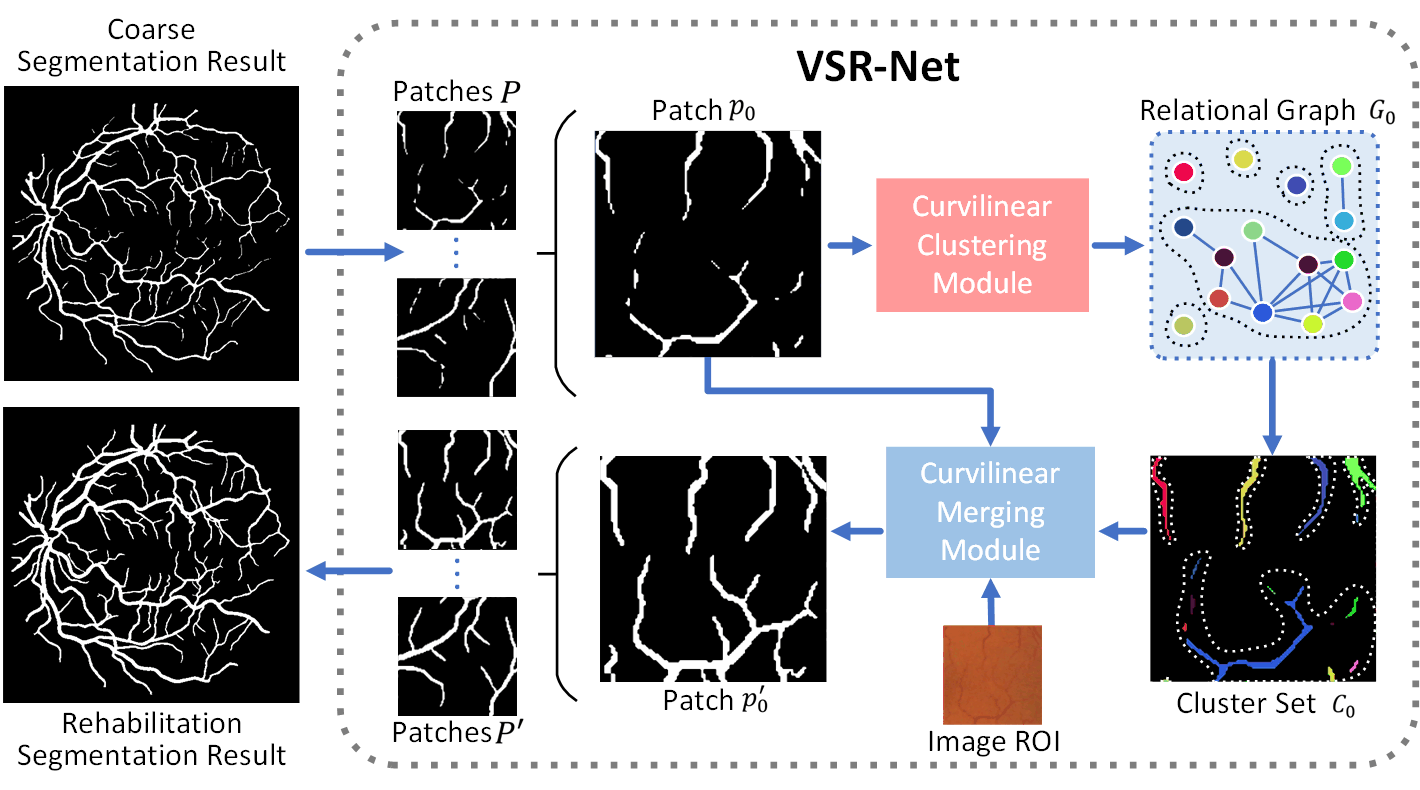}}
	\caption{Overview of the VSR-Net. (1) the coarse segmentation result is divided into individual patches $P$ and then we input them into the Curvilinear Clustering Module (CCM). (2) CCM constructs a relationship graph $G$ based on the vessel-like subsection ruptures in the input patch and then generates cluster set $C$. (3) the Curvilinear Merging Module (CMM) rehabilitate the subsection ruptures in the input patch according to cluster set $C$, and combines the rehabilitation results of all patches to obtain the rehabilitation segmentation result of the entire image.}
	\label{fig2}
\end{figure*}

\subsection{Vessel-like Structures Segmentation}

Over the years, scholars have achieved significant progress in vessel-like structure segmentation tasks with 2D and 3D semantic segmentation methods. For \emph{2D Vessel-like Structures Segmentation}: Gu et al. proposed CENet~\cite{CENet} for blood vessel-like structure segmentation (e.g., fundus blood vessels \cite{DRIVE}) by spatial pyramid pooling. Mou et al. presented a CS2Net\cite{CS2Net} to segment curve structures, which applied a dual self-attention module to enhance the representational ability of deep networks. Similarly, other studies \cite{Sklcon} incorporated attention modules and multi-channel feature fusion strategies into deep network-based semantic segmentation methods for highlighting salient features in curvilinear segmentation tasks. In \emph{3D Vessel-like Structures Segmentation}: In contrast to 2D vessel-like structures, 3D vessel-like structures exhibit indescribable intricacies in terms of textural characteristics and elongation of curvilinear \cite{ATM, ERNet}. Cicek et al. proposed a 3D U-Net \cite{3DUNet} by prompting U-Net \cite{UNet} in 3D medical image segmentation tasks with 3D convolution operations. Hatamizadeh et al. \cite{UNETR}proposed a UNETR, which effectively took advantage of a Transformer encoder and a CNN decoder to improve segmentation results. Apart from deep network architecture designs, other studies have focused on multi-scale feature representation enhancement and rigorous loss constraints for enhancing small vessel-like structure segmentation. Although existing methods have achieved considerable vessel-like structure segmentation results, they had difficulty in preserving the structural continuity of the segmented curvilinear, bringing subsection rupture problems. Deep network-based refinement segmentation methods have been considered as a possible approach to address the subsection rupture problem in vessel-like structure segmentation tasks by refining coarse segmentation results.

\subsection{Deep Network-based Refinement Segmentation}
Deep network-based refinement segmentation aims to generate a more precise depiction of objects or areas within an image by refining coarse segmentation results. Jha et al. \cite{DoubleUNet} proposed a Double UNet architecture to generate coarse and refinement segmentation results with two UNets, respectively. Following this two-stage refinement segmentation paradigm, Zhu et al. \cite{RFNet} proposed an RFNet to fuse coarse and refined segmentation results by refining input image patches. Li et al. \cite{AGLN} proposed an attention-directed global enhancement module and local refinement network (AGLN) for aggregating the global semantic information. Chen et al. proposed an RRCNet \cite{RRCNet}, composed of SegNet with a deep supervision module, a missed detection residual network, and a false detection residual network. Nham et al. \cite{Nham2023Re}  proposed a two-stage refinement segmentation framework, which adopted EffcientNet as the feature encoder and applied Tversky loss to constrain coarse and fine segmentation results. Moreover, Zhang et al. \cite{DARN} presented the DARN to refine coarse segmentation results on 3D images by attention mechanisms. Xia et al. \cite{ERNet} proposed an ER-Net based on an encoder-decoder structure, which applied a feature selection module to adaptively select discriminative features from the encoder and decoder for improving edge segmentation results. Liang et al. \cite{CPM} proposed a refinement segmentation model to mitigate biases inherent in coarse segmentation outputs via a context pooling module (CPM). These methods improve segmentation results by designing various feature learning modules and loss constraints at the cost of increasing the complexity of deep networks. In particular, they ignore the integrity of vessel-like structures or the particularity of the subsection rupture problem, causing them to address the subsection rupture problem poorly.

\subsection{Graph Convolutional Network in Segmentation}
Recently, GCNs\cite{GCNSu} have widely been applied to tackle semantic segmentation tasks and have achieved significant results. The current research directions of GCNs are roughly divided into two categories. (1) Feature attribute learning module: Gan et al. \cite{GCN_C1} proposed a class-based dynamic graph convolution (CDGC) module to adaptively propagate information and perform graph reasoning between pixels of the same category to improve the segmentation results. Li et al. \cite{GCN_C2} proposed a dynamic channel graph convolutional network edge enhancement (DE-DCGCN-EE) based on dual encoders, aiming to fuse edge and spatial features. Other studies have converted categories, attributes, features, and such on into graph information. Subsequently, they utilize GCN for feature extraction and induction, enhancing the segmentation completeness. (2) Interactive segmentation: Ling et al. \cite{GCN_I1} introduced the Curve-GCN that applied a graph structure to segment contours or surfaces, and then used GCN to evolve the curve to achieve interactive segmentation results iteratively. Kim et al.~\cite{GCN_I3} proposed the Split-GCN based on the polygon method and self-attention mechanism, enabling the vertices of the interactive boundary to move to the object boundary more accurately by providing direction information. Other studies also defined the edge points of objects as nodes within a graph, then conducted feature extraction through GCN for predicting the position offsets of graph nodes to encompass the object edges. GCN has advanced advantages in graph structure data feature extraction and spatial correlation analysis, which has rarely been applied to address the subsection rupture problem in vessel-like structure segmentation tasks. Hence, in contrast to previous methods, we are the first to refine the subsection rupture problem as the rehabilitation task instead of the segmentation task, aiming to take advantage of GCNs and refinement segmentation methods to rehabilitate subsection ruptures in coarse vessel-like structure segmentation results with negligible overhead.

\subsection{Semantic Segmentation Calibration}

Confidence calibration is to predict probability by estimating representative of true correctness likelihood, which is crucial in computer-aided medical diagnosis. Over the years, several works have been proposed to tackle calibration in semantic segmentation. Wang et al. \cite{Unc3} systematically studied the calibration of semantic segmentation methods and proposed a selective scaling method to improve the predicted segmentation confidence. Alireza et al.~\cite{9130729} proposed model ensembling to improve confidence calibration of deep network-based semantic segmentation methods in medical image segmentation tasks. Ding et al. \cite{ding2021local} presented a local temperature scaling method to extend temperature scaling to each pixel. Other methods, such as adversarial training, domain shift, MC-dropout, and stochastic processes, are also introduced to enhance confidence calibration in semantic segmentation tasks. Unlike previous methods, we attempt to improve confidence calibration of vessel-like structure segmentation results from the structure rehabilitation perspective.

\begin{figure*}[!thb]
	\centerline{\includegraphics[width=0.75\linewidth]{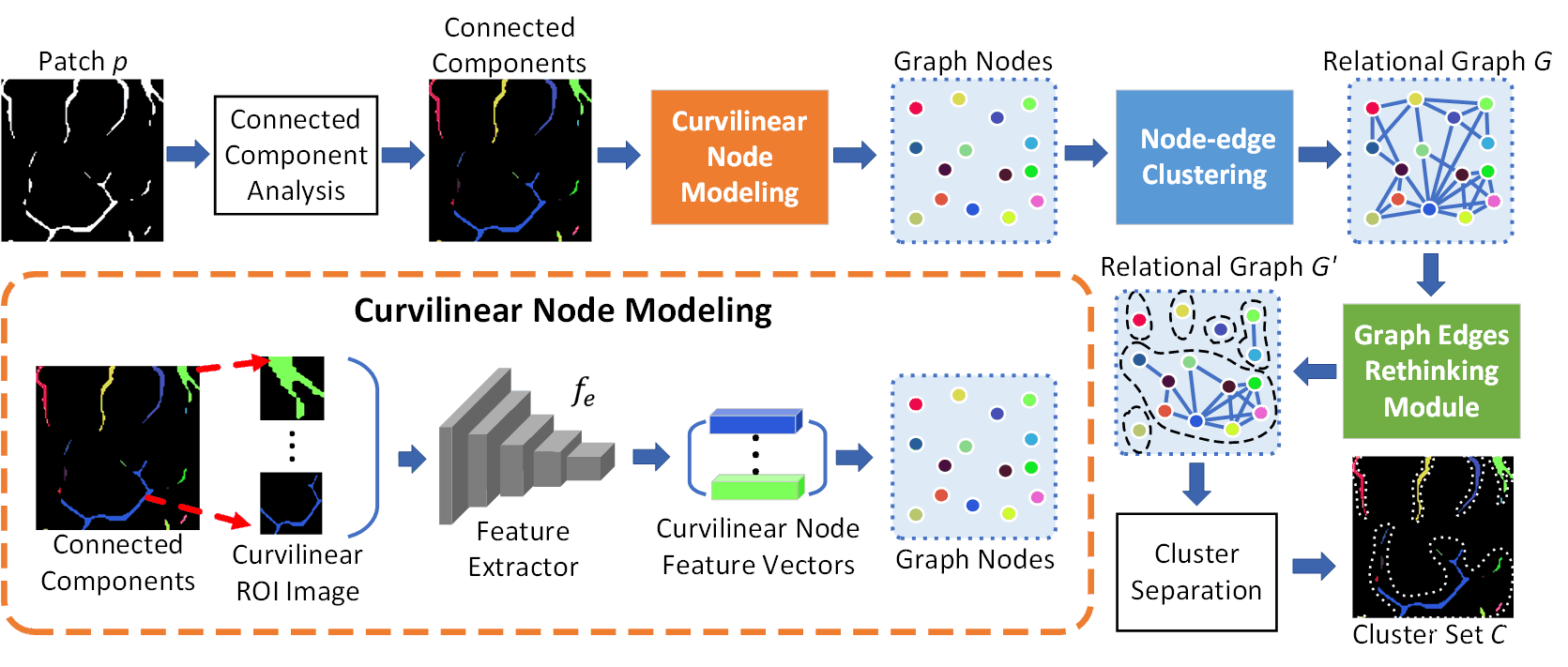}}
	\caption{Overview of the Curvilinear Clustering Module (CCM) process. CCM first models the subsection rupture of the patch image $p$ into the nodes of the graph through curvilinear node modeling; Then, it clusters the edges of the nodes through the Node-edge clustering algorithm to obtain the relationship graph $G$; Finally, the refined relationship graph $G'$ is obtained by pruning the edges without a connection relationship through the graph edge rethinking module, and dividing them through cluster separation to obtain the cluster set $C$.}
	\label{fig3}
\end{figure*}

\section{Method}
\label{sec:method}
\subsection{Overview of VSR-Net}

In this paper, we introduce a novel Vessel-like Structure Rehabilitation Network (VSR-Net) to rehabilitate coarse vessel-like structure segmentation results caused by subsection ruptures problem, as shown in Fig.\ref{fig2}.
Our VSR-Net comprises two main modules: Curvilinear Clustering Module (CCM) and Curvilinear Merging Module (CMM). As shown in Fig.\ref{fig2}, the general flowchart of VSR-Net is described as follows: 

\begin{itemize}
\item \textbf{Preprocessing}: For coarse segmentation results (such as from U-Net\cite{UNet}), using a sliding window to segment it into a set of patches.
\item \textbf{Subsection Ruptures Clustering}: Enter each patch into the CCM module, which generates a relationship graph $G$ and then extracts cluster set $C$ from it. According to Fig.\ref{fig2}, the subsection ruptures with white dashed lines in each cluster should be rehabilitate as a complete vessel-like structure object.
\item \textbf{Subsection Ruptures Rehabilitation}: The CMM rehabilitates the subsection ruptures in the patch based on cluster set $C$, patch image and its corresponding image ROI, and finally combines all the rehabilitation results to obtain the rehabilitation result of the entire image.
\end{itemize}

Considering the integrity of vessel-like structures, most subsection ruptures can be viewed as part of some vessel-like structures from a global perspective. Thus, inputting the entire coarse segmentation image into VSR-Net directly, may cause CMM to aggregate all subsection ruptures into a single cluster. The cluster results contradict our expectations. To address this problem, at the preparation stage, we use the sliding window strategy to divide the input image into patches $P=[p_{0}, p_{1},...,p_{n}]$ (as shown in Fig.\ref{fig2}) and then input them into VSR-Net. In this paper, we set the sliding window size to 1/16 of the size of the original image in the 2D images, e.g., the length and width are 1/4 of the original image. For the 3D image, the size of the sliding window is 64$\times$64$\times$64. The following sections will introduce CCM and CMM in detail.

\subsection{Curvilinear Clustering Module}

Fig.\ref{fig3} shows an overview of the CCM. When each patch image is input into CCM, the connected components in the patch image is first obtained through connected component analysis. As shown in Fig.\ref{fig3}, each subsection rupture is marked with a different color to represent a connected component. The next process of CCM is as follows:
\begin{itemize}
\item \textbf{Curvilinear Node Modeling}: It maps the connected components into graph nodes. Each node represents a connected component, that is, a subsection rupture.

\item \textbf{Node-edge Clustering}: Based on the obtained graph nodes, the Node-edge clustering algorithm is applied to cluster graph nodes, and then we merge clustering graph nodes to construct relationship graph $G$.
\item \textbf{Graph Edge Rethinking}: Following Node-edge Clustering, this paper uses the graph edges rethinking module to extract graph features and prune the edges without connection relationships, aiming to obtain the refined relationship graph $G'$. According to the connection status of the nodes in $G'$, all nodes are split into several clusters to obtain the cluster set $C$ with cluster separation.
\end{itemize}

\subsubsection{Curvilinear Node Modeling} 
Each subsection rupture in the patch image can be viewed as a node in the graph, and the minimum distance between different subsection ruptures can be used to represent the weight of the edges for corresponding nodes. Therefore, the spatial interconnection correlation of subsection ruptures can be converted into a relationship graph $G$. The subsection ruptures of this relationship graph for vessel-like structure can be described as a tuple $G(N, E)$. Where $N$ signifies the node feature set of the graph via a matrix with dimensions $|N| \ast L$, where $|N|$ represents the number of graph nodes, and $L$ represents the dimension of the node feature vector. Meanwhile, $E$ represents the edge set within the graph. Note that constructing the relationship graph $G$ involves converting vessel-like structure subsection ruptures into graphical representations. 

In order to obtain the node feature set $N$ of the relationship graph $G$, we adopt ResNet-18 as a feature extractor $f_{e}$ to encode each curvilinear ROI image into feature vectors with 512. Specifically, we disentangle the curvilinear ROI image of each subsection rupture from the patch image by computing the minimum enclosing rectangle (MER) area of each connected component, then eliminating other irrelevant subsection ruptures. This operation ensures each curvilinear ROI image only contains the corresponding single subsection rupture object (as shown in Fig.\ref{fig3}, curvilinear node modeling for the orange dashed line area). The process to obtain the feature vector of each subsection rupture is formulated as follows:
\begin{equation}
\begin{split}
	N &= \{n_1, n_2, ..., n_{|N|}\}\\
      &=\{f_{e}(v_1), f_{e}(v_2), ..., f_{e}(v_{|N|})\}, \label{eq:1}
\end{split}
\end{equation}
\label{eq1}
Where $\{n_1, n_2, ..., n_{|N|}\}$ are the node features with length $L=512$, each node feature $n$ carries out feature extractor on the curvilinear ROI image $v$ through the feature extractor $f_{e}$. The feature vector of each node represents the feature value of the corresponding subsection rupture.

\begin{figure}[b]
\centerline{\includegraphics[width=0.85\columnwidth]{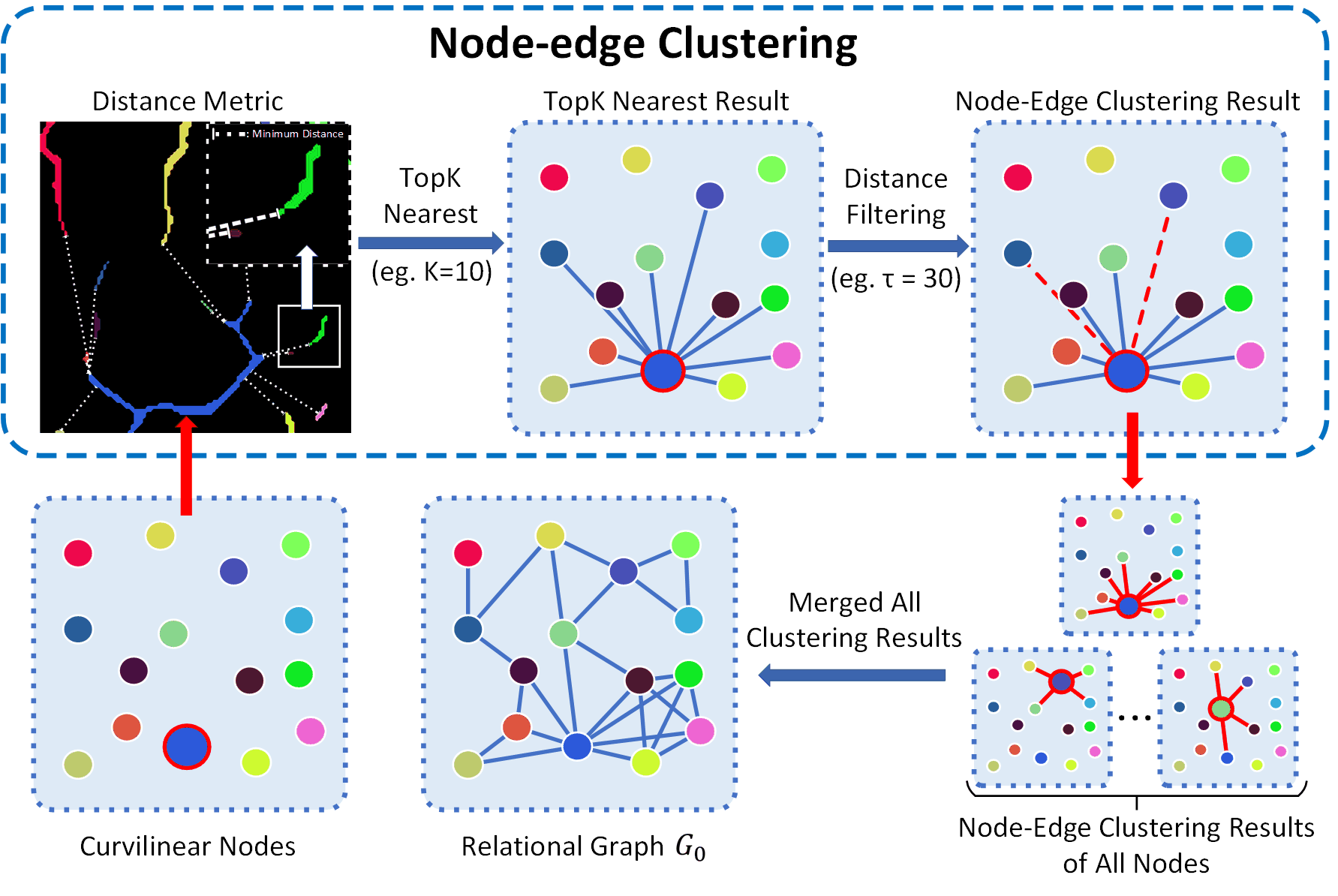}}
	\caption{Overview of Node-edge Clustering process. For each curvilinear node clustering needs processing distance metric, Top-$K$ nearest neighbor search and distance threshold filtering respectively. Then, all the clustering results are combined to get the edge set $E$ of the relationship graph $G$.}
	\label{fig4}
\end{figure}

\subsubsection{Node-edge Clustering} 
Next, adopting the Top-$K$ nearest subsection ruptures as neighbor nodes, as shown in Fig.\ref{fig4}. Here, we take $K=10$ as an example, the target node (red node) connects the ten closest nodes based on the Top-$K$ nearest results. Subsequently, setting a distance threshold $\tau$, then the edges with a distance greater than $\tau$ will be trimmed; finally, the remaining edges will be used as the useful edges of the node. As shown in Fig.\ref{fig4}, taking $\tau=30$ as an example, the target node (red node) prunes the two nearest neighbors whose shortest distance exceeds the threshold $\tau$. Based on this, the edges of each node are clustered by the node-edge clustering method. Then, the clustering results of each node are combined to generate the complete graph edge set $E$. If the edge between two nodes is not pruned, it indicates that the edge is valid; that is, there is a connection between the two nodes. After determining the graph's complete graph edge set $E$, the edges' weight is normalized by relationship graph $G$. We set the total edge number to $Z$ and the weight $E_i$ for the $i$ edge as:
\begin{equation}
	D_{A,B} =  \mathop{min}\limits_{a \in A, b \in B} f_{dis}(a,b) , 
\end{equation}
where $f_{dis}$ represents the \textbf{euclidean distance metric} function, the $a$ and $b$ are points of subsection ruptures $A$ and $B$, respectively, indicating the shortest distance between the two subsection ruptures. 

Next, Adopting the Top-$K$ nearest subsection ruptures as neighbour nodes. As shown in Fig.\ref{fig4}, take $K=10$ as an example, the target node (red node) connects the ten closest nodes in the TopK nearest result. Subsequently, set a distance threshold $\tau$, and the edges with a distance greater than $\tau$ will be trimmed, and the remaining edges will be used as the effective edges of the node. As shown in Fig.\ref{fig4}, take $\tau=30$ as an example, the target node (red node) prunes the two nearest neighbors whose shortest distance exceeds the threshold $\tau$. Based on this, the edges of each node are clustered by node-edge clustering method. Then, the clustering results of each node are combined to get the complete graph edge set $E$. If the edge between two nodes is not pruned, it indicates that the edge is valid, indicating that there is a connection between the two nodes. Algorithm 1 summarizes the clustering details of node-edge clustering. After determining the graph's complete graph edge set $E$, the edges' weight is normalized by relationship graph $G$. Set that there are $Z$ edges in total, and the weight $E_i$ of the $i$ edge is:
\begin{equation}
	E_{i} = \frac{e^{\frac{1}{D_{i}}}}{\sum_{z=1}^Z e^{\frac{1}{D_{z}}}},\label{eq3}
\end{equation}
where, $D_{i}$ represents the original minimum distance of the $i$ edge. The Eq.(3) adopts the Softmax function to normalize the weight of the edge. If the shortest distance between two nodes larger, the weight of the edge between them should be smaller. Thus, the relationship graph $G$ is composed of the set of graph edges set $E=\{e_1,e_2,...,e_P\}$ and the node feature set $N=\{n_1, n_2, ..., n_{|N|}\}$. 

\begin{algorithm}[t]
	\caption{Node-edges Search Algorithm} 
	\hspace*{0.02in} {\bf Input:} 
	Vascular image set $\{v\}$. Initialised parameters of neighbour matrix set $\{w\}$. Hyperparameter of maximum neighbours number $K$ and distance threshold $\tau$. \\
	\hspace*{0.02in} {\bf Output:} 
	The parameters of the neighbour matrix set $\{w\}$.
	\begin{algorithmic}[1]
		\For{$i$ in range( len($v$ ))}
		\State Initialized temp list, $T$=[  ].
		\For{$j$ in range( len($v$ ))}
		\If{$i \neq j$} 
		\State Compute the minimum  distance $D_{v_i, v_j}$.
		\State Append the $D_{v_i, v_j}$ to $T$.
		\EndIf 
		\EndFor
		\State Search the first $K$ minimum distance value indexes in $T$, get the indexes set $r$, the len($r$) $\leq K$.
		\For{$j$ in range( len($v$ ))}
		\If{$D_{v_i, v_j} < \tau$ and $j \in r$}. 
		\State $w_{i,j} = T_{j}$, connection.
		\Else
		\State $w_{i,j} = \infty$, non-connection.
		\EndIf 
		\EndFor
		\EndFor
		\State \textbf{return} $\{w\}$
	\end{algorithmic}
\end{algorithm}

\subsubsection{Graph Edges Rethinking Module}
Graph Edge Rethinking Module (GERM) considers the interconnection relationship among the subsection ruptures, represented by each node in the relationship graph $G$. GERM also prunes the edges without connection relationships in $G$. The feature extraction process in the GERM is implemented by GCN, which is formulated as follows:
\begin{equation}
	h_{i}^{l+1} = \sigma(\sum_{j \in N_i }  \frac{1}{c_{ij}} h_{j}^{l} w^l + b^l),
\end{equation}
\label{eq4}
where $h_{j}^{l}$ is the feature of node $i$ at the $j$ layer, $\sigma$ is the LeakReLU function. ${c_{ij}}$ is a normalized factor, $h_{j}^{l}$ and $b^l$ are trainable weights and bias of the $l$ layer. $N_i$ is the set of neighboring nodes of node $i$, including its own nodes. Eq.(4)describes the operation flow of layer $l+1$ in GCN. 

Moreover, to dynamically adjust the relative importance of neighbours of each node, we introduce the attention mechanism \cite{GAT} in GCN by calculating the attention coefficient of the current node and its neighbors. Then, the node features are weighted and summed according to the calculated attention coefficient. The attention mechanism in the GCN layer is computed as follows:
\begin{equation}
	h_{i}^{l+1} = \sigma(\frac{1}{K} \sum_{k=1}^K \sum_{j \in N_i }  \frac{1}{c_{ij}} \alpha_{ij}^k h_{j}^{l} W^{l,k} + b^l),
 \label{eq6}
\end{equation}
where $ \alpha_{ij}^k $ is the normalised attention coefficients  computed by the $k$-th attention mechanism. After the graph feature representation extraction of the GCN layer, the end of the GERM is the FC layer for classifying edge categories. Specifically, the categories of each edge in the refined relationship graph $G'$ is predicted. The two-node features connected by each edge are spliced and then input to the FC Layer to get the prediction result in $ y_{E_i} = \Theta(F_{fc}([h_a||h_b])) $, $\Theta$ is Sigmod function. When the predicted result is 0, this edge will be deleted. In contrast, this edge will remain when the result is 1. 

\begin{figure*}[t]
	\centerline{\includegraphics[width=0.75\linewidth]{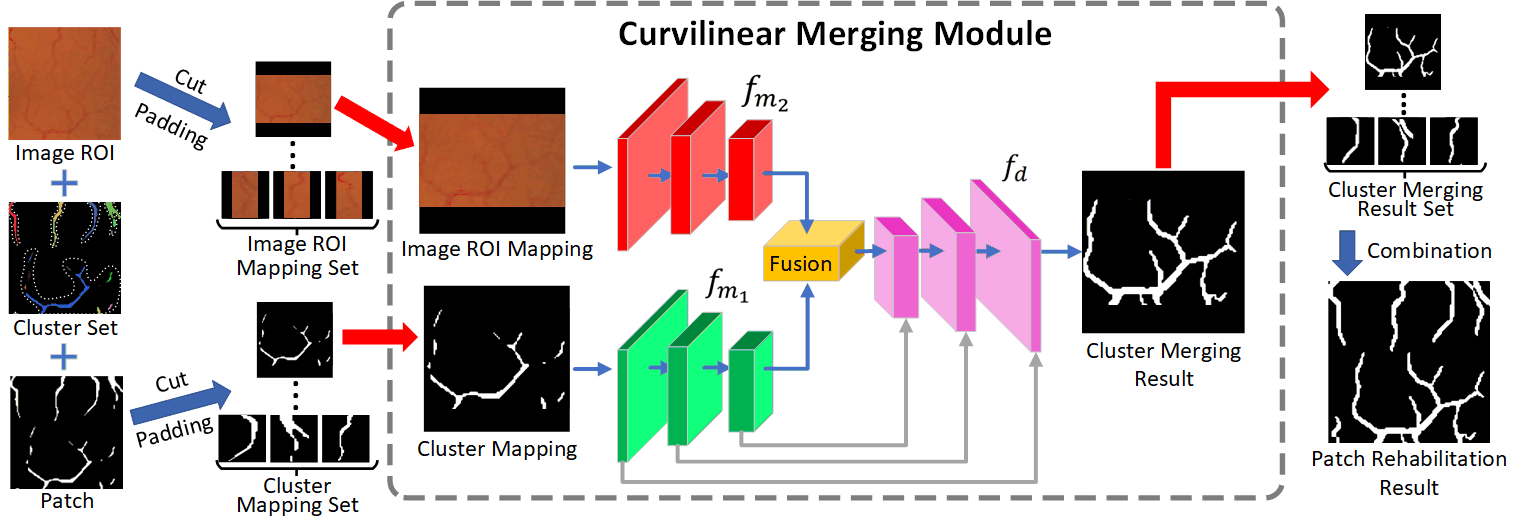}}
	\caption{Overview of the Curvilinear Merging Module (CMM) process. The CMM cuts and pads the mapping sets of the image ROI and Patch based on the cluster set. Subsequently, these sets are paired and input into the CMM module to derive cluster merging results. Finally, the entirety of cluster merging results is consolidated into the patch rehabilitation results.}
	\label{fig5-1}
\end{figure*}

Finally, as for the design of GERM loss, in addition to the classification loss of edge prediction, regularisation is also adopted to enrich the diversity of multi-head attention learning:
\begin{equation}
	L_{gcn} = \frac{1}{T} \sum_{t=1}^T L_{bce}(y^{\prime}_{E_t}, y_{E_t}) + L_{r}(O), \label{eq7}
\end{equation}
\begin{equation}
	L_{r}(O) = -\frac{1}{L*H^2} \sum_{l=1}^L \sum_{i=1}^H \sum_{j=1}^H \frac{O_{l,i} \cdot O_{l,j}}{||O_{l,i}|| ||O_{l,j}||}, \label{eq8}
\end{equation}
where $L_{bce}$ is the binary cross-entropy loss as edge classification loss. $y_{E_t}$ is the predicted result of GERM, and $y^{\prime}_{E_i}$ is the label of the edge. $L_{r}(O)$ is attention regularization, $O_{l,i}$ represents the output of $i$-th GCN attention at layer $l$. Specifically, Eq.(7) calculates the cosine similarity $cos(\cdot)$ between the same layer attention outputs pair through the normalised dot product. This regularisation directly applies to the outputs of each attention head by maximising the difference among them. 

\subsection{Curvilinear Merging Module} 
According the process of CCM, we obtain the cluster set $C$. To merge the subsection ruptures of each cluster into a complete subsection, we propose the curvilinear merging module (CMM). The process of merging the subsection ruptures into an entire subsection via the proposed CMM, as shown in Fig.\ref{fig5-1}. In each cluster, CMM initially captures the minimum enclosing rectangle area around the subsection ruptures. Subsequently, it eliminates any other subsection ruptures located in this area that do not belong to this cluster.
CMM cuts and pads the image ROI and patch respectively, for producing the mapping sets corresponding to the clusters. These two mapping sets are fed into two feature encoder modules, denoted as $f_{m_1}$ and $f_{m_2}$, performing feature extraction on the ROI area $R$ and the original image $I$, respectively. The feature representations obtained from these two encoder modules are then integrated with the fusion step. Subsequently, fused feature representations are passed through the decoder, represented as $f_{d}$, to generate the cluster merging result.

The encoder and decoder modules in CMM are implemented based on the Light UNet architecture \cite{LUNet}. To enhance vessel-like structure rehabilitation performance, we delete two sets of convolutional layers and downsampling layers at the high stage of both encoder and decoder, resulting in a downsampling ratio change from 32 to 8. This adjustment aims to preserve more shallow features. Additionally, skip connection mechanisms have been introduced between the encoder and decoder to transmit texture and shape information across different stages. Finally, all the cluster merging results are mapped according to their relative positions and combined into patch-based vessel-like structure rehabilitation results.

Considering curvilinears only occupy a small area of the patch image, the pixel cross-entropy loss and Dice loss cannot describe the subtle differences in the vessel-like segmentation well. The CMM loss is proposed as follows:
\begin{equation}
	L_{cmm} = L_{ce}(R^{'}, \hat{R}^{'}) + L_{pl}(R^{'}, \hat{R}^{'}) + L_{cr}(R^{'}),
 \label{eq9}
\end{equation}
where $R'$ is the result of the revised merged model, and $\hat{R'}$ is the ground truth. $ L_{ce} $ is the cross-entropy loss responsible for guiding at the beginning of model training. $L_{pl}$ is the polygon line loss, which mainly measures the polyline similarity between $R'$ and $\hat{R'}$:
\begin{equation}
	L_{pl} = \frac{1}{|Q|} \sum_{i=1}^{|Q|} \mathop{min}\limits_{q \in Q} ||s_i - q||_2 + \frac{1}{|S|} \sum_{j=1}^{|S|} \mathop{min}\limits_{s \in S} ||s - q_j||_2,
\end{equation}
\label{eq10}
where $S$ and $Q$ are pixel points sets of subsections in $R'$ and $\hat{R'}$ respectively.

$L_{pl}$ measures the distance difference of subsection pixels between $ S $ and $ Q $. In the CMM, we expect all subsection ruptures in $R'$ to be merged into a more complete subsection. When the number of connected components in $R'$ is greater than 1, indicating that there are subsection ruptures or subsections that still have not been merged. To address this problem, we propose connected region loss function t$ L_{cr} $:$ L_{cr} = 1 -  \frac{1}{C}$, where $C$ is the number of connected components of $\hat{R'}$. When the number of connected components of $\hat{R'}$ decreases, the loss will also decrease, encouraging the CMM to segment the subsections together to obtain promising vessel-like structure rehabilitation results.

\section{EXPERIMENTS}
We first introduce datasets in Sec.\ref{sec4.1}. Then, implementation details and evaluation metrics are listed in Sec.\ref{sec4.2}. Followed by ablation studies and performance comparisons with SOTA methods in Sec.\ref{sec4.3} and \ref{sec4.4}. Finally, we provide quantitative analysis to demonstrate the robustness of our method in Sec.\ref{sec4.5}.

\begin{table}[h]
	\caption{Information from five public datasets}\label{tab1}
	\centering
    \renewcommand\arraystretch{1.4}
	\begin{tabular}{ccccc}
		\hline
		Dataset &  Training-test &  Object &Modality \\
		\hline
		DRIVE &  20-20 & Fundus vessels & Fundus images \\
		OCTA-500 & 240-60 & Fundus vessels & OCTA \\
		CORN-1 & 1176-340 & Corneal nerve & CCM images \\
        \hline
		ATM & 200-99 & Pulmonary trachea& CT \\
		MIDAS-I & 30-20 & Intracranial artery & MRI \\
		\hline
	\end{tabular}
	\vspace{-5pt}
    \label{table:table1}
\end{table}

\subsection{Datasets}
\label{sec4.1}
In this paper, we adopt five 2D/3D medical image datasets to verify the effectiveness of our method. The 2D medical image datasets include the DRIVE, OCTA-500,  and CORN-1. The 3D medical image datasets include the ATM and MIDAS-I, as listed in Table \ref{table:table1}.

\subsubsection{2D vessel-like segmentation datasets}\ 

\textbf{DRIVE}\cite{DRIVE}\footnote{http://www.isi.uu.nl/Research/Databases/DRIVE/}. It contains 40 pairs of fundus images and corresponding labels of vessel-like object segmentation. The size of each image in the dataset is 565$\times$584. 

\textbf{OCTA-500}\cite{OCTA500}\footnote{https://ieee-dataport.org/open-access/octa-500}. It contains OCTA images of the eyes of 300 patients with 6mm × 6mm FOV. Only one image for each subject's eye, ensuring diversity and avoiding similar data. The size of each image in the dataset is 400$\times$400.

\textbf{CORN-1}\cite{CORN1}\footnote{https://imed.nimte.ac.cn/CORN.html}. It contains 1516 CCM images of subbasal corneal epithelium using a Heidelberg Retina Tomograph equipped with a Rostock Cornea Module (HRT-III) microscope. Each image has a resolution of 384 × 384 pixels covering a FOV of 384$\times$384 $\mu^{m2}$ with manual annotation. 

\subsubsection{3D vessel-like segmentation datasets}\

\textbf{ATM}\cite{ATM}\footnote{https://atm22.grand-challenge.org/}. The dataset contains 299 CT scans from multi-sites. Three radiologists carefully label the airway tree structures with more than five years of professional experience. Each chest CT scan consisted of various slices, ranging from 157 to 1125, with a slice thickness of 0.450–1.000 mm. The axial size of all slices is 512 × 512 pixels with a spatial resolution of 0.500–0.919 mm. 

\textbf{MIDAS-I}\cite{ERNet}\footnote{https://public.kitware.com/Wiki/TubeTK/Data}. It is a subset of MIDAS, including 50 manual annotations of intracranial arterial in MRA. MIDAS dataset is a public cerebrovascular dataset that contains a total of 100 MRA volumes that were acquired from healthy volunteers aged 18 to 60+ years. A core region containing intracranial arteries measuring 224 × 208 × 64 was cropped from each raw data to minimise interference with other brain regions.

\subsection{Experimental Settings}
\label{sec4.2}
\subsubsection{Implementation Details}

We implement our VSR-Net and comparable methods with the Pytorch platform. Stochastic gradient descent (SGD) method is adopted as the optimizer for updating parameters with the momentum (0.99) and the weight decay (0.0001). We set the training epochs to 100. The batch size of CCM and CMM was set to 4 and 10, respectively. The learning rates of CCM and CMM are $10^{-3}$ and $10^{-4}$, respectively. The batch size and learning rate for other segmentation networks are 4 and $10^{-4}$ in training. Standard data augmentation methods are applied during training, such as random scaling (from 0.5 to 2.1) and random flipping. It is worth noting that the data size input to the model during inference is the same as during the training phase; thus, we can execute all patches for the same image in parallel. All experiments are performed using two GPUs (Tesla V100).

\subsubsection{Evaluation Metrics}

In this paper, we evaluate the general performance of our method from three views qualitatively and quantitatively: segmentation metrics, vessel-like structure rehabilitation metrics, and confidence calibration metrics.

\begin{itemize}
\item Segmentation metrics. Three commonly used evaluation measures are adopted: Pixel Accuracy ($PA$), Dice coefficient ($Dice$), and Jaccard coefficient ($Jacc$).
\item Vessel-like structure rehabilitation metrics. According to the clinical requirements \cite{Amir2017,zhao2020automated, van2020vascular,chua2020retinal}, we use Vascular Bifurcation Number ($VBN$), Fractal Dimension ($FD$), and Vascular Tortuosity ($VT$) to vessel-like morphology rehabilitation results.
\item Confidence calibration metrics. The widely accepted expected calibration error ($ECE$) is utilized to measure the calibration of the network. 
\end{itemize}


\subsection{Ablation Studies}
\label{sec4.3}

In this section, we perform ablation experiments to validate the effectiveness of CCM and CMM modules in VSR-Net, including: 1) hyperparameter selection of the CCM module, and 2) evaluation of the effects of each component in the CMM module. In the preparation stage, we adopt UNet \cite{UNet} as the baseline to generate coarse segmentation results based on the DRIVE dataset.

\begin{figure*}[!]
	\centerline{\includegraphics[width=0.7\linewidth]{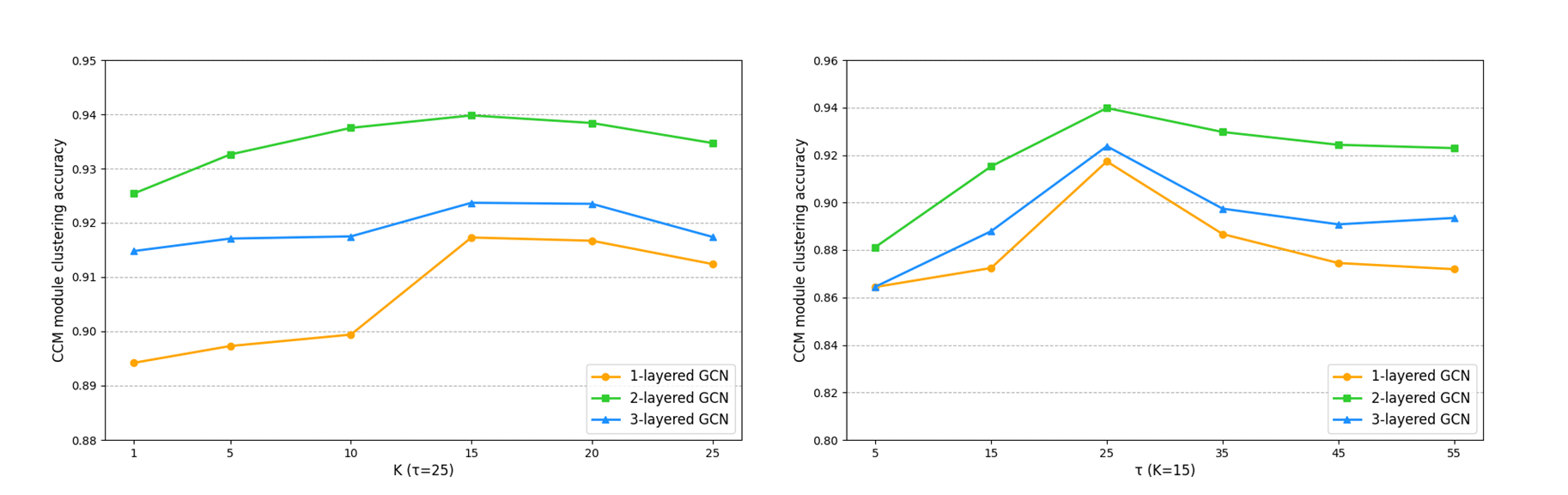}}
	\caption{The performance of CMM when node nearest neighbours TopK $K$, the distance threshold $\tau$, and the number of GCN layers taking different values on DRIVE \cite{DRIVE} test set. When $K$=15, $\tau$=25, and GCN layers=2, the CCM module builds the highest-quality performance.}
	\label{fig5}
\end{figure*}

\subsubsection{Hyperparameter Selection in CCM}

We test the effects of different node nearest neighbors top-K ($K$) value settings on the CCM module, pruning distance threshold ($\tau$) and the depth $l$ of GCN layers. Fig.\ref{fig5} offers the performance comparisons of different hyperparameter selections: $\tau$ and $l$. According to Fig.\ref{fig5}(Left), the CCM achieves the best performance with $K=15$ when $\tau=25$ pixels is fixed. Fig.\ref{fig5}(Right)  presents the performance comparisons of different $\tau$ values in the CMM when $K=15$ is fixed. It can be observed that CCM performs better with $\tau=25$ than other $\tau$ value settings. In addition, we also see that GCN obtains the best performance when the depth $l$ of GCN is 2 than other depth of layer setting, as shown in Fig.\ref{fig5}.
Overall, the hyperparameter selections of $K$, $\tau$, and $l$ are crucial for the performance of CCM in clustering the graph $G$. When $K=15$, $\tau-25$, and $l=2$, CCM obtains the highest performance,  and we will adopt these settings for the following experiments.

\begin{table}[h]
	\caption{Ablation study of CMM on DRIVE \cite{DRIVE} test set}
	\centering
	\begin{tabular}{ccc|ccc}
		\hline
		$\rightarrow_{skip}$ & $L_{pl}$ & $L_{cd}$ & $PA$ & $Dice$ & $Jaccard$\\
		\hline
		&  &  & 0.9504 & 0.8415 & 0.7090 \\
		\checkmark &  &  & 0.9567 & 0.8580 & 0.7322 \\
		\checkmark & \checkmark &  & 0.9592 & 0.8657 & 0.7435 \\
		\checkmark & \checkmark & \checkmark & \textbf{0.9621} & \textbf{0.8790} & \textbf{0.7629} \\
		\hline
	\end{tabular}
	\vspace{-5pt}
     \label{table:table2}
\end{table}

\subsubsection{Effects of each compoment in CMM} 
Table~ \ref{table:table2}lists the results of each component in the CMM: skip connection ($\rightarrow_{skip}$), polygon line loss ($L_{pl}$), and connected domain loss ($L_{cd}$). While each component brings meaningful performance improvements, the combination of these three components further boosts the performance, demonstrating that CMM is capable of rehabilitating subsection ruptures. 

\begin{table*}

	\caption{Results comparisons of 
    state-of-the-art methods and them with  VSR-Net in terms of segmentation metrics, Vessel-like structure rehabilitation metrics, and condifidence calibration metrics on the DRIVE \cite{DRIVE} and OCTA-500 \cite{OCTA500} test sets.}\label{tab3}
	\centering
	\renewcommand\arraystretch{1.5}
	\setlength{\tabcolsep}{3.5pt}{} 
    \begin{threeparttable}
	\begin{tabular}{c *{7}{c} | *{7}{c}}
        \hline
		\multirow{2}{*}{Methods} & \multicolumn{7}{c}{DRIVE} & \multicolumn{7}{c}{OCTA-500} \\
        \cmidrule(r){2-15}

		& $PA\uparrow$ & $Dice\uparrow$ & $Jacc\uparrow$ & $VBN\downarrow$ & $FD\downarrow$ & $VT\downarrow$ &  $ECE\downarrow$ & $PA\uparrow$ & $Dice\uparrow$ & $Jacc\uparrow$ & $VBN\downarrow$ & $FD\downarrow$ & $VT\downarrow$ & $ECE\downarrow$ \\
        \hline
		
		DoubleUNet\cite{DoubleUNet} &
        0.9567 & 0.8504 & 0.7242 & 0.3671 & 0.5891 & 0.6643 & 0.0678 &
		0.9755 & 0.8590 & 0.7504 & 0.2907 & 0.4894 & 0.4243 & 0.0790 \\

		RFNet\cite{RFNet} & 
        0.9569 & 0.8538 & 0.7257 & 0.3321 & 0.5849 & 0.6417 & 0.0771 &
		0.9799 & 0.8643 & 0.7527 & 0.2671 & 0.4757 & 0.3963 & 0.1195 \\
		
		AGLN\cite{AGLN} & 
        0.9573 & 0.8538 & 0.7356 & 0.3175 & 0.5764 & 0.5993 & 0.0506 &
		0.9824 & 0.8654 & 0.7533 & 0.2867 & 0.4076 & 0.3581 & 0.0929 \\

		RRCNet\cite{RRCNet} & 
        0.9593 & 0.8645 & 0.7428 & 0.2918 & 0.5426 & 0.5953 & 0.0679 &
		0.9844 & 0.8668 & 0.7610 & 0.2171 & 0.4056 & 0.3286 & 0.0736 \\
		
		EN-UNet\cite{Nham2023Re} & 
        0.9604 & 0.8678 & 0.7546 & 0.2861 & 0.5401 & 0.5404 & 0.0622 &
		0.9856 & 0.8718 & 0.7647 & 0.2160 & 0.3388 & 0.3109 & 0.0824\\

        \hline
        U-Net \cite{UNet} & 
        0.9504 & 0.8415 & 0.7090 & 0.4844 & 0.6179 & 0.7155 & 0.0786 &
		0.9733 & 0.8539 & 0.7453 & 0.3604 & 0.5580 & 0.4806 & 0.1058\\

		VSR-Net$^\dag$ & 
        \textbf{0.9621} & \textbf{0.8790} & \textbf{0.7629} & \textbf{0.1743} & \textbf{0.5039} & \textbf{0.4690} & \textbf{0.0362} &
		
		\textbf{0.9881} & \textbf{0.8767} & \textbf{0.7809} & \textbf{0.1574} & \textbf{0.2508} & \textbf{0.2038} & \textbf{0.0435}\\
		
        \hline
	\end{tabular}

    \begin{tablenotes}
        \footnotesize
        \item[*] VSR-Net$ ^\dag $ is the refinement segmentation result obtained after using VSR-Net on the coarse segmentation result of U-Net \cite{UNet}.
    \end{tablenotes}
     \end{threeparttable}

    \label{table:table3}
\end{table*}

\begin{table}
    \scriptsize
	\caption{Results comparisons of state-of-the-art methods and them with VSR-Net in terms of segmentation metrics, Vessel-like structure rehabilitation metrics, and condifidence calibration metrics on the CORN-1 \cite{CORN1} test set.}
	\centering
	\renewcommand\arraystretch{1.5}
    \setlength{\tabcolsep}{3pt}
    \begin{threeparttable}
	\begin{tabular}{c|*{7}{c}}
		\hline
		Model & $PA\uparrow$ & $Dice\uparrow$ & $Jacc\uparrow$ & $VBN\downarrow$ & $FD\downarrow$ & $VT\downarrow$ &  $ECE$\\
		\hline

        DoubleUNet\cite{DoubleUNet} & 
        0.9492 & 0.8585 & 0.7507 & 0.5120 & 0.5644 & 0.5971 & 0.0907 \\

        RFNet\cite{RFNet} & 
        0.9528 & 0.8600 & 0.7531 & 0.5117 & 0.5346 & 0.5853 & 0.0522 \\

        AGLN\cite{AGLN} & 
        0.9531 & 0.8604 & 0.7548 & 0.4094 & 0.5302 & 0.5407 & 0.0724 \\

        RRCNet\cite{RRCNet} & 
        0.9562 & 0.8634 & 0.7571 & 0.3943 & 0.5292 & 0.4743 & 0.0653 \\

        EN-UNet\cite{Nham2023Re} & 
        0.9566 & 0.8690 & 0.7604 & 0.3448 & 0.5273 & 0.4724 & 0.0983 \\

        \hline
        U-Net \cite{UNet} & 
        0.9433 & 0.8539 & 0.7453 & 0.6662 & 0.5787 & 0.7167 & 0.0901 \\
  
		VSR-Net$^\dag$ & 
        \textbf{0.9657} & \textbf{0.8724} & \textbf{0.7678} & \textbf{0.2316} & \textbf{0.4964} & \textbf{0.3805} & \textbf{0.0388} \\
		
		\hline
	\end{tabular}

    \begin{tablenotes}
        \footnotesize
        \item[*] VSR-Net$ ^\dag $ is the refinement segmentation result obtained after using VSR-Net on the coarse segmentation result of U-Net \cite{UNet}.
    \end{tablenotes}
     \end{threeparttable}

    \label{table:table4}
\end{table}

\begin{figure}[!]
	\centerline{\includegraphics[width=0.8\columnwidth]{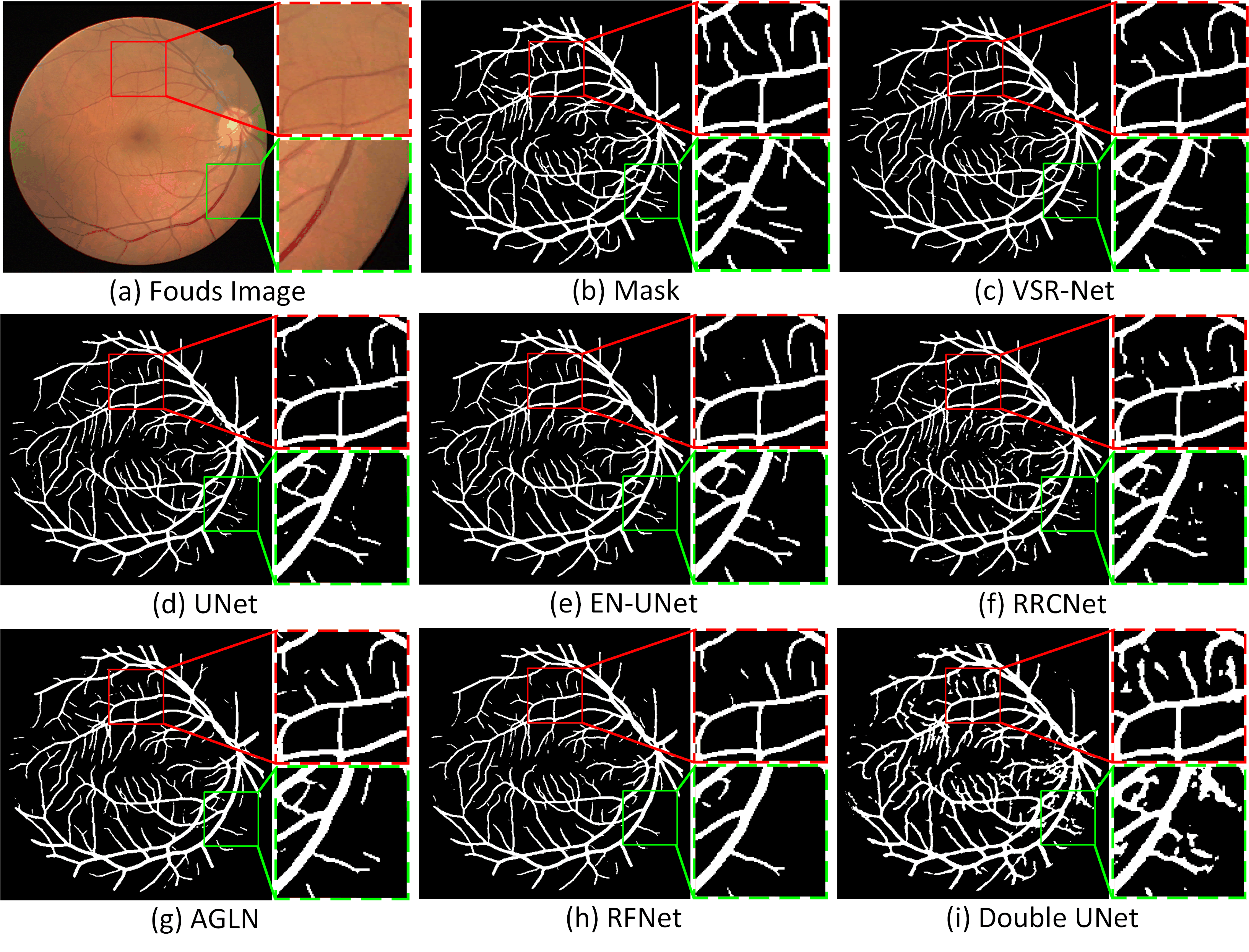}}
	\caption{Visual comparison between vessel-like structure segmentation results by previous methods and vessel-like structure rehabilitation results by VSR-Net in a representative sample from the DRIVE\cite{DRIVE} test set.}
	\label{fig6}
\end{figure}
\begin{figure}[!]
	\centerline{\includegraphics[width=0.8\columnwidth]{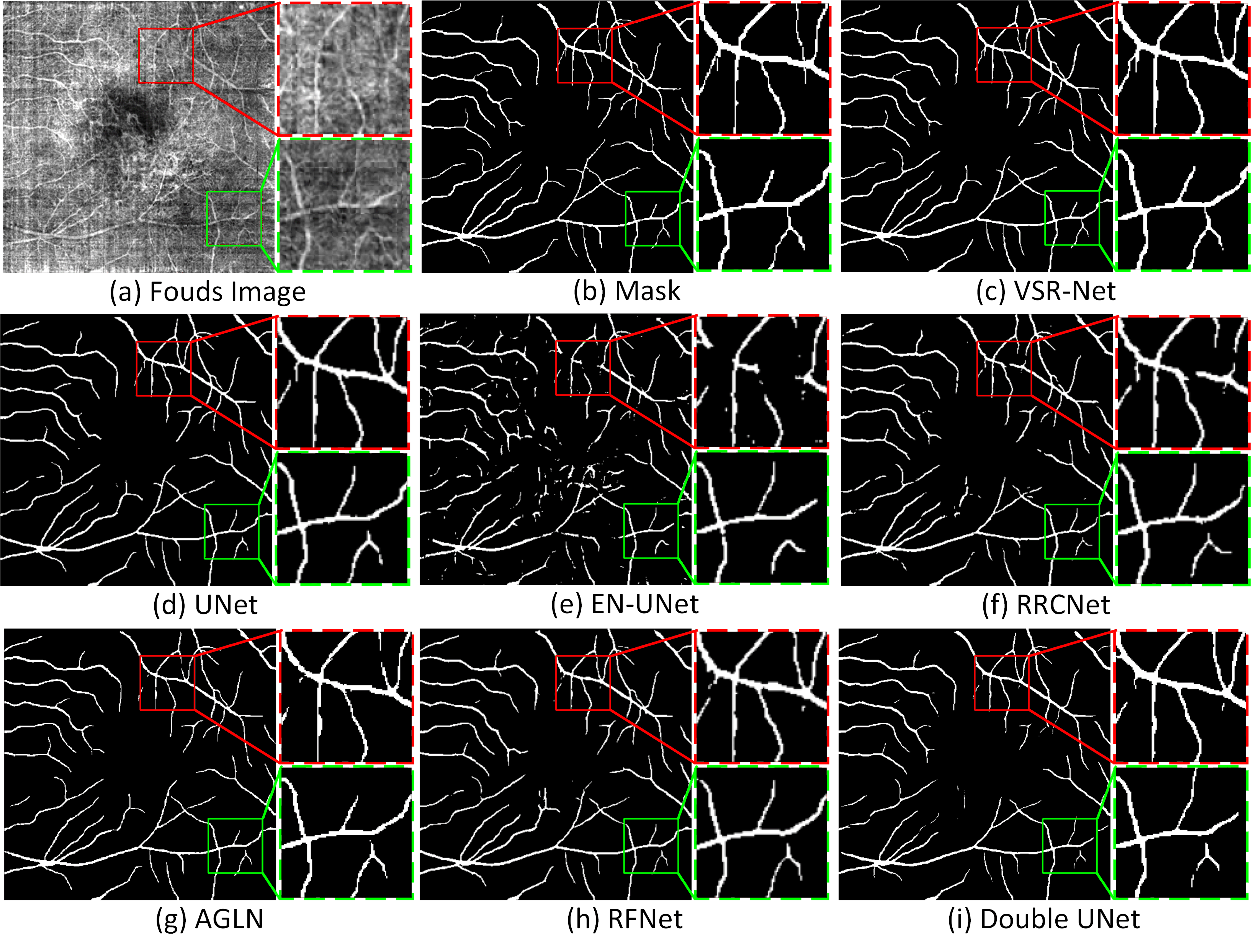}}
	\caption{Visual comparison between vessel-like structure segmentation results by previous methods and vessel-like structure rehabilitation results by VSR-Net in a representative sample from the OCTA-500\cite{OCTA500} test set.}
	\label{fig7}
\end{figure}
\begin{figure}[!]
	\centerline{\includegraphics[width=0.8\columnwidth]{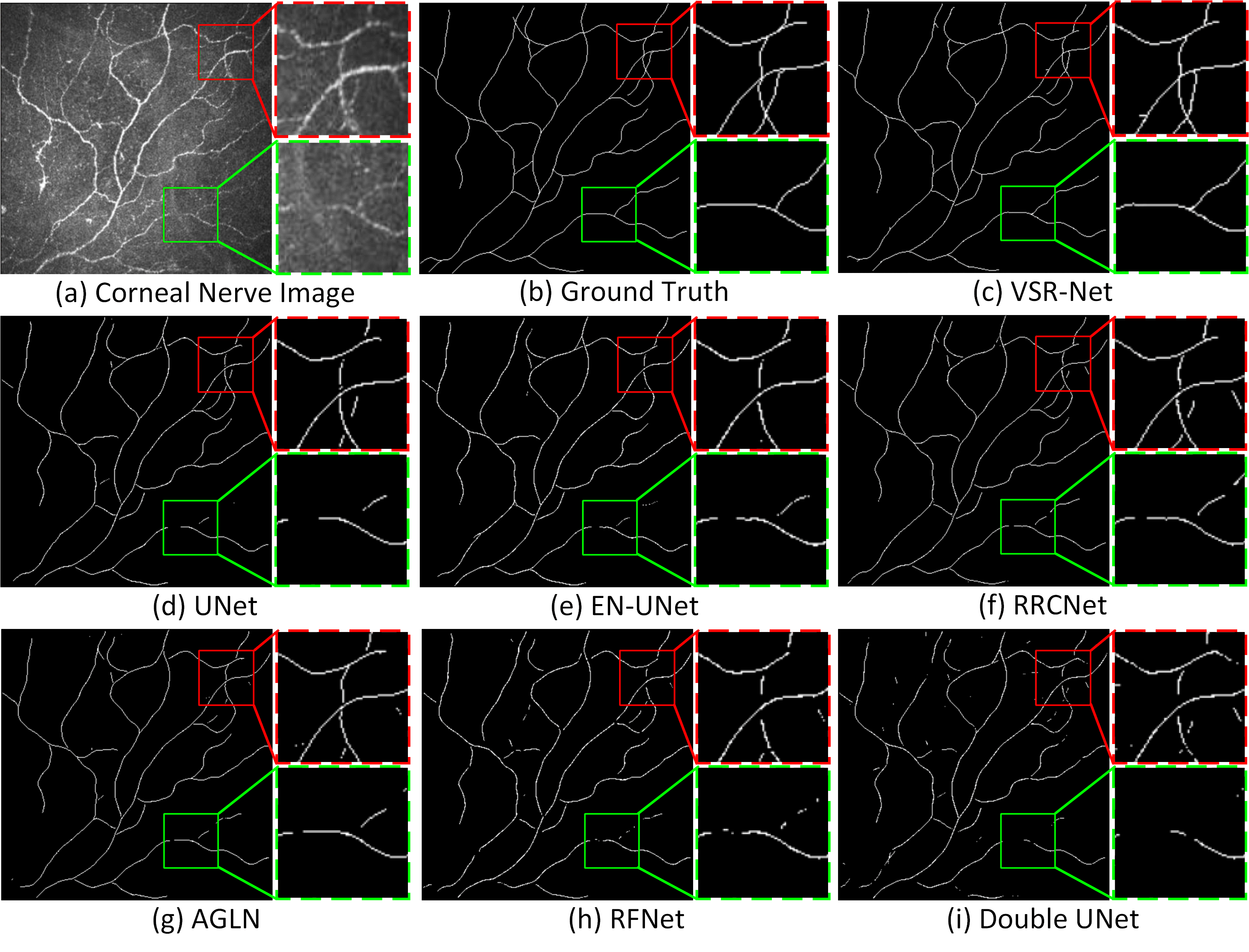}}
	\caption{Visual comparison between vessel-like structure segmentation results by previous methods and vessel-like structure rehabilitation results by VSR-Net in a representative sample from the CORN-1\cite{CORN1} test set.}
	\label{fig8}
\end{figure}

\subsection{Comparisons with State-of-the-Arts}
\label{sec4.4}
This section compares our VSR-Net with five SOTA refinement segmentation methods on five 2D/3D vessel-like structure segmentation datasets. (1) 2D refinement segmentation methods: Double-UNet~\cite{DoubleUNet}, RFNet~\cite{RFNet}, AGLN~\cite{AGLN}, RRCNet~\cite{RRCNet}, and EN-UNet\cite{Nham2023Re}. (2) 2D refinement segmentation methods: DARN~\cite{DARN}, EfficientSegNet~\cite{EfficientSegNet}, ERNet~\cite{ERNet}, CPM~\cite{CPM} and UR-CarA-Net\cite{URCarA}.

For our VSR-Net, We utilized 2D U-Net\cite{UNet} and 3D U-Net\cite{3DUNet} to generate coarse vessel-like structure segmentation results. Then, VSR-Net dynamically rehabilitates the subsection ruptures based on these coarse segmentation results. 

\subsubsection{Performance Comparison on 2D Vessel-like Structure Datasets}
Table\ref{table:table3} and Table\ref{table:table4} offers the comparison results of our VSR-Net and competitive methods in terms of vessel-like structure segmentation results, vessel-like structure rehabilitation results, and ECE results on the 2D vessel-like structure datasets. We conclude as follows: (1) VSR-Net performs better than other SOTA refinement segmentation methods on the DRIVE\cite{DRIVE} test set, benefiting from the well-directed structure rehabilitation with graph clustering. (2) We observe that VSR-Net outperforms the method proposed by EN-UNet \cite{Nham2023Re}, which also focuses on boundary-aware of segmentation task, improving the PA by 0.0017, the Dice by 0.0112, and the Jaccard by 0.0083 respectively. (3) For vessel-like structure rehabilitation results comparison, our method significantly reduces the VBN by absolute over 0.1118 through comparisons to other SOTA methods. (4) Our method also achieves the best results of FD and VT. Regarding confidence calibration, we also observe that the ECE of VSR-Net is lower than other SOTA methods and decreases the ECE  in the range of 0.0409 to 0.0144. Similarly, VSR-Net also achieves the best results of seven evaluation metrics on OCTA-500 \cite{OCTA500} and CORN-1 \cite{CORN1} test sets. Overall, the experimental results in Table\ref{table:table3}and Table \ref{table:table4} show the superiority of VSR-Net in rehabilitating subsection ruptures and improving the confidence calibration of deep networks with graph clustering over existing deep network-based refinement segmentation methods.

Next, we provide 2D structure rehabilitation visualization comparisons of three different vessel-like structures, as listed in Fig. \ref{fig6}, Fig. \ref{fig7} and Fig. \ref{fig8}, where (a) is the input image; (b) is manually annotated ground truth; (c)-(i) are VSR-Net, U-Net, Nham et al., RRCNet, AGLN, RFNet and Double U-Net, respectively. According to Fig.~ \ref{fig6}- Fig. \ref{fig8}, we obtain the following conclusions: (1) there are many subsection ruptures in the coarse vessel-like structure results of U-Net compared with the ground truth; (2) Deep network-based refinement segmentation methods such as AGLN and Nham et al. can improve vessel-like structure segmentation results, but they still cannot solve the subsection ruptures problem well without considering relationships of subsection ruptures. (3) As shown in the two enlarged areas in three vessel-like structures, VSR-Net can effectively rehabilitate the subsection ruptures and lead them to merge. Moreover, some of the poorly segmented microvessels obtained more accurate boundaries after rehabilitation by VSR-Net; (4) The visual structure rehabilitation result comparisons indicate our VSR-Net not only obtains promising segmentation results but also efficiently rehabilitates subsection ruptures in vessel-like structures, agreeing with our expectation.

\begin{table*}
	\scriptsize
	\caption{Results comparisons of state-of-the-art methods and them with VSR-Net in terms of segmentation metrics, vessel-like structure rehabilitation metrics, and condifidence calibration metrics on the ATM \cite{ATM} and MIDAS-I \cite{ERNet} test sets.}
	\centering
	\renewcommand\arraystretch{1.5}
	\setlength{\tabcolsep}{3.5pt}{} 
    \begin{threeparttable}
	\begin{tabular}{c *{7}{c} | *{7}{c}}
        \hline     
		\multirow{2}{*}{Methods} & \multicolumn{7}{c}{ATM} & \multicolumn{7}{c}{MIDAS-I} \\
  
        \cmidrule(r){2-15}

		& $PA\uparrow$ & $Dice\uparrow$ & $Jacc\uparrow$ & $VBN\downarrow$ & $FD\downarrow$ & $VT\downarrow$ &  $ECE\downarrow$ & $PA\uparrow$ & $Dice\uparrow$ & $Jacc\uparrow$ & $VBN\downarrow$ & $FD\downarrow$ & $VT\downarrow$ & $ECE\downarrow$ \\
  
        \hline

        DARN\cite{DARN} &
        0.9262 & 0.7253 & 0.5956 & 0.2606 & 0.3328 & 0.4830  & 0.0893 &
		0.9333 & 0.8519 & 0.7720 & 0.5405 & 0.5119 & 0.5853  & 0.0770 \\

        EfficientSegNet\cite{EfficientSegNet} & 
        0.9275 & 0.7254 & 0.5971 & 0.2623 & 0.3121 & 0.4822 & 0.0797 &
		0.9346 & 0.8543 & 0.7723 & 0.5362 & 0.5001 & 0.5412 & 0.0547 \\

        ERNet\cite{ERNet} & 
        0.9283 & 0.7293 & 0.5993 & 0.2772 & 0.3016 & 0.4726 & 0.0701 &
		0.9361 & 0.8548 & 0.7744 & 0.5191 & 0.4665 & 0.5214 & 0.0466 \\

        CPM\cite{CPM} & 
        0.9283 & 0.7309 & 0.6004 & 0.2623 & 0.2988 & 0.4480 & 0.0605 &
		0.9371 & 0.8548 & 0.7758 & 0.5023 & 0.4088 & 0.5021 & 0.0463 \\

        UR-CarA-Net\cite{URCarA} & 
        0.9293 & 0.7355 & 0.6032 & 0.2606 & 0.2907 & 0.3732 & 0.0866 &
		0.9373 & 0.8551 & 0.7780 & 0.4932 & 0.3894 & 0.4874 & 0.0547 \\

        \hline
        3D U-Net\cite{3DUNet} &
        0.9217 & 0.7209 & 0.5886 & 0.4418 & 0.3637 & 0.5307 & 0.0752 &
		0.9315 & 0.8458 & 0.7688 & 0.6245 & 0.5709 & 0.6841 & 0.0677 \\
        
		VSR-Net$^\dag$ & 
        \textbf{0.9359} & \textbf{0.7519} & \textbf{0.6127} & \textbf{0.2038} & \textbf{0.2739} & \textbf{0.3113} & \textbf{0.0475} &
        
        \textbf{0.9397} & \textbf{0.8586} & \textbf{0.7835} & \textbf{0.3955} & \textbf{0.3909} & \textbf{0.4196} & \textbf{0.0251} \\
		\hline

        \hline
	\end{tabular}

    \begin{tablenotes}
        \footnotesize
        \item[*] VSR-Net$ ^\dag $ is the refinement segmentation result obtained after using VSR-Net on the coarse segmentation result of 3D U-Net \cite{3DUNet}.
    \end{tablenotes}
 \end{threeparttable}
 \label{table:table5}
\end{table*}

\subsubsection{Performance Comparison on 3D Vessel-like Structure Segmentation Datasets}

Table\ref{table:table5} offers the comparison results of our VSR-Net and competitive methods on the 3D vessel-like structure datasets. It can be observed that our VSR-Net consistently performs better than comparable 3D refinement segmentation methods. For example, compared to UR-CarA-Net, VSR-Net reduces three vessel-like structure rehabilitation metrics on the ATM test set by 0.0568, 0.0168 and 0.0619, respectively. Moreover, the gain of VSR-Net on dice score is larger than 0.0164, and the reduction of ECE is over 0.0130 on the ATM test set. Furthermore, we also obtain similar conclusions on the MIDAS-I dataset, demonstrating the generalization ability and effectiveness of VSR-Net from three evaluation views.

\begin{figure}
	\centerline{\includegraphics[width=0.75\columnwidth]{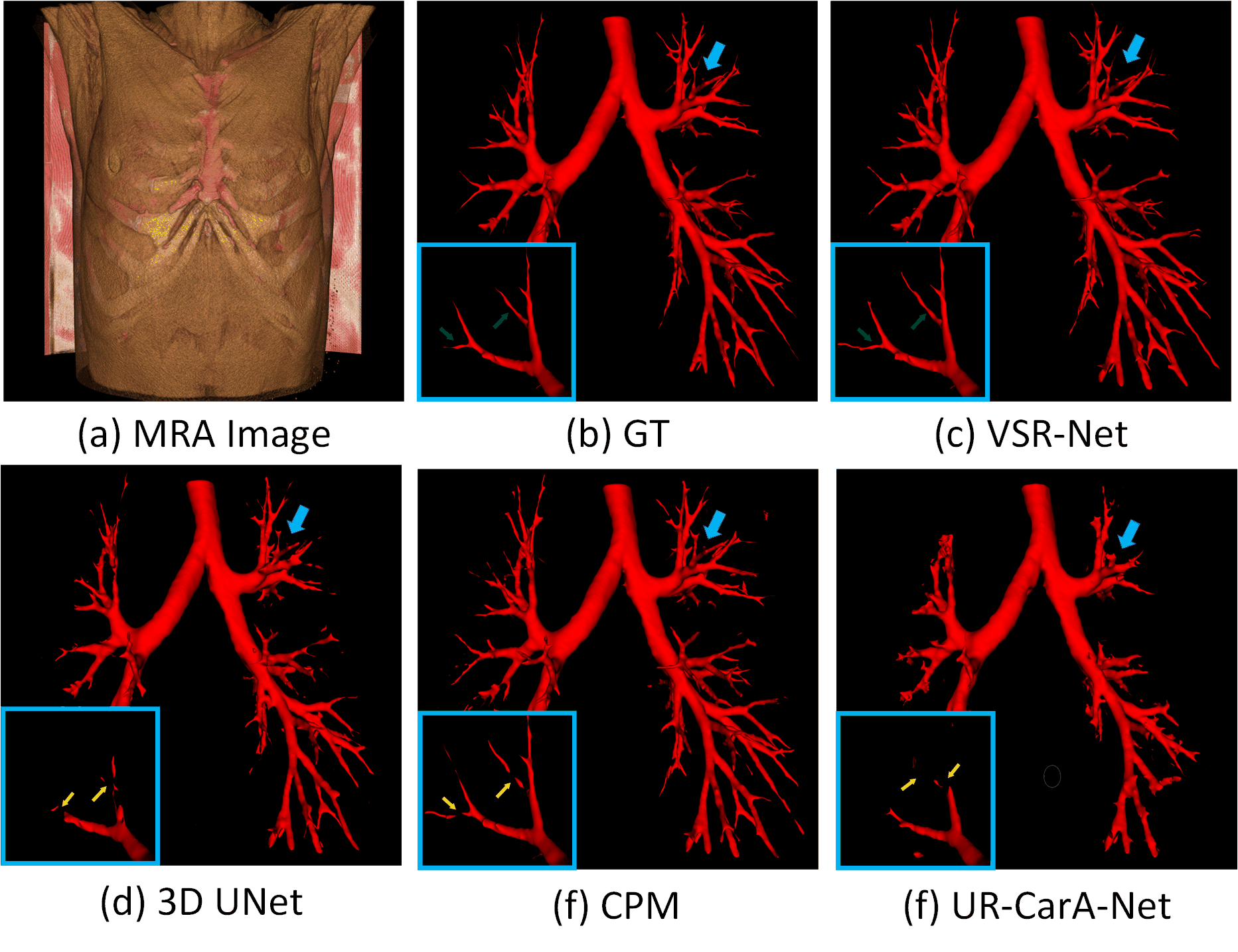}}
	\caption{Visual comparison between vessel-like structure segmentation results by previous methods and vessel-like structure rehabilitation results by VSR-Net in a representative sample from the ATM\cite{ATM} test set.}
	\label{fig9}
\end{figure}

\begin{figure}
	\centerline{\includegraphics[width=0.75\columnwidth]{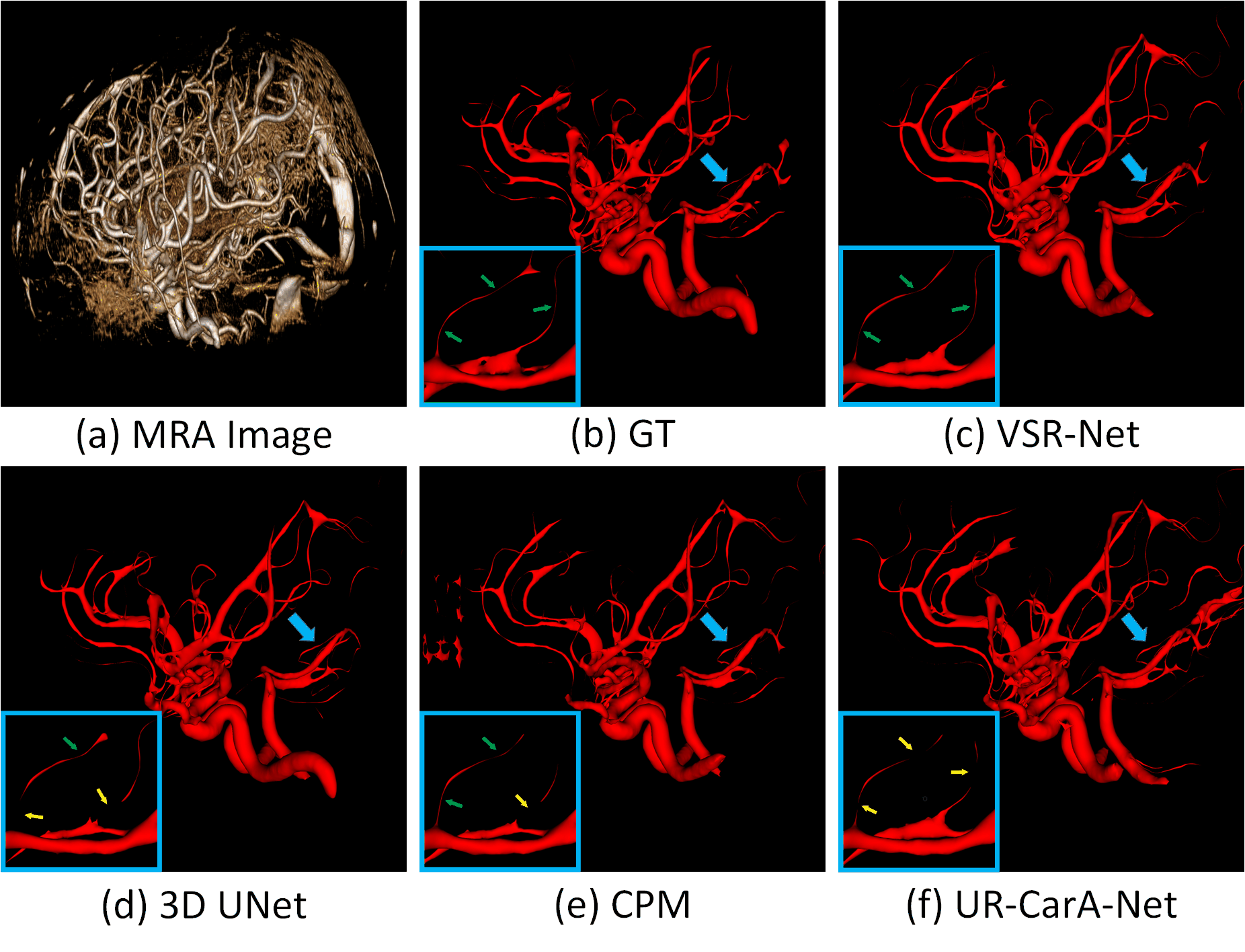}}
	\caption{Visual comparison between vessel-like structure segmentation results by previous methods and vessel-like structure rehabilitation results by VSR-Net in a representative sample from the MIDAS-I\cite{ERNet} test set.}
	\label{fig10}
\end{figure}

Fig. \ref{fig9} and Fig. \ref{fig10} present 3D structure rehabilitation visualization comparisons of two vessel-like structures, where (a) shows the input image; (b) is the manually annotated ground truth; (c)-(f) are VSR-Net, 3D U-Net, CPM and UR-CAr-Net, respectively. From t Fig. \ref{fig9} and Fig. \ref{fig10}, we can see as follows: (1) Some end connecting parts of the 3D vessel-like structures relatively slender, and comparative deep network-based refinement segmentation methods produce rupture subsections more frequently in these places. (2) We observe that the segmentation results of VSR-Net are more complete and malleable on the terminal micro-trachea than that of GT by comparing the magnification areas (b) and (c) in Fig. \ref{fig9}.
(3) Similar to 2D vessel-like structure rehabilitation visualizations, the results show that VSR-Net achieves promising performance in 3D structure rehabilitation and structure segmentation with graph clustering, proving the generalization and superiority of our method through comparisons to advanced deep network-based refinement segmentation methods.


\subsection{Generalization and Reliability Analysis}
\label{sec4.5}

To further prove generalization and reliability of our method on 2D and 3D vessel-like structure rehabilitation tasks, we conduct extensive experiments on the DRIVE \cite{DRIVE} and ATM \cite{ATM} datasets, respectively. First, we utilize commonly used 2D/3D vessel-like structure segmentation baselines to generate coarse vessel-like structure segmentation results. Here, 2D vessel-like structure segmentation baselines include UNet \cite{UNet}, CENet \cite{CENet}, CS $^2$ -Net \cite{CS2Net} and SKelcon-Net \cite{Sklcon}, and 3D vessel-like structure segmentation baselines include 3D UNet \cite{3DUNet}, nnUNet \cite{nnUNet}, UNet \cite{UNETR} and Swin UNETR \cite{SUNETR}. Finally, this paper makes comparisons between original vessel-like structure segmentation baselines and them with VSR-Net from three different dimensions: segmentation metrics, vessel-like structure rehabilitation metrics, and confidence calibration metrics. 

\begin{table*}
	\caption{Results comparisons of baseline methods integrated with VSR-Net in terms of segmentation metrics, vessel-like structure rehabilitation metrics, and condifidence calibration metrics on the DRIVE\cite{DRIVE} test set.}\label{tab2}
	\centering
	\renewcommand\arraystretch{1.5}
	\setlength{\tabcolsep}{3.5pt}{} 
	\begin{tabular}{c*{7}{c} }
        \hline   
		Methods & $PA\uparrow$ & $Dice\uparrow$ & $Jacc\uparrow$ & $VBN\downarrow$ & $FD\downarrow$ & $VT\downarrow$ & $ECE\downarrow$ \\
	    \hline
		UNet \cite{UNet} & 0.9504 & 0.8415 & 0.7090 & 0.4844 & 0.6179 & 0.7155 & 0.0786\\
		
		+VSR-Net & \textbf{0.9621} & \textbf{0.8790} & \textbf{0.7629} & \textbf{0.1743} & \textbf{0.5039} & \textbf{0.4690} & \textbf{0.0362}\\
		\hline
  
		 CENet \cite{CENet} & 0.9569 & 0.8483 & 0.7184 & 0.5161 & 0.6417 & 0.7660 & 0.0603\\
		
		+VSR-Net & \textbf{0.9650} & \textbf{0.8864} & \textbf{0.7606} & \textbf{0.1769} & \textbf{0.3834} & \textbf{0.4724} & \textbf{0.3379}\\
		\hline
  
		CS$^2$-Net \cite{CS2Net} & 0.9396 & 0.8249 & 0.7021 & 0.5979 & 0.5893 & 0.6584& 0.0846\\
		
		+VSR-Net & \textbf{0.9596} & \textbf{0.8561} & \textbf{0.7460} & \textbf{0.1835} & \textbf{0.4005} & \textbf{0.0332} & \textbf{0.0460} \\
		\hline
  
		SKelcon-Net \cite{Sklcon} & 0.9675 & 0.8567 & 0.7205 & 0.5650 & 0.6523 & 0.6509 & 0.0592\\
		
		+VSR-Net & \textbf{0.9782} & \textbf{0.8769} & \textbf{0.7652} & \textbf{0.1635} & \textbf{0.5233} & \textbf{0.4097} & \textbf{0.0283}\\
	    \hline
	\end{tabular}
   \label{table:table6}
\end{table*}

\begin{table*}
	\caption{Results comparisons of baseline methods integrated with VSR-Net in terms of segmentation metrics, vessel-like structure rehabilitation metrics, and condifidence calibration metrics on the ATM\cite{ATM} test set.}\label{tab2}
	\centering
	\renewcommand\arraystretch{1.5}
	\setlength{\tabcolsep}{3.5pt}{} 
	\begin{tabular}{c*{7}{c} }
		\hline
		Methods & $PA\uparrow$ & $Dice\uparrow$ & $Jacc\uparrow$ & $VBN\downarrow$ & $FD\downarrow$ & $VT\downarrow$ & $ECE\downarrow$  \\
		\hline
  
		3D UNet \cite{3DUNet} & 0.9217 & 0.7209 & 0.5886 & 0.4418 & 0.3637 & 0.5307 & 0.0752\\
		
		+VSR-Net & \textbf{0.9359} & \textbf{0.7519} & \textbf{0.6127} & \textbf{0.2038} & \textbf{0.2739} & \textbf{0.3113} & \textbf{0.0475}\\
		\hline
  
		nnUNet \cite{nnUNet} & 0.9356 & 0.7685 & 0.6150 & 0.4721 & 0.3803 & 0.4817 & 0.0820 \\
		
		+VSR-Net & \textbf{0.9428} & \textbf{0.8060} & \textbf{0.6699} & \textbf{0.2634} & \textbf{0.2063} & \textbf{0.2607} & \textbf{0.0569} \\
		\hline
  
		UNET \cite{UNETR} & 0.9324 & 0.7538 & 0.6017 & 0.4639 & 0.2782 & 0.4748 & 0.0735\\
		
		+VSR-Net & \textbf{0.9435} & \textbf{0.7974} & \textbf{0.6442} & \textbf{0.1932} & \textbf{0.1662} & \textbf{0.2452} & \textbf{0.0337} \\
		\hline
  
		Swin UNETR \cite{SUNETR} & 0.9416 & 0.7772 & 0.6163 & 0.4281 & 0.2939 & 0.4370 & 0.0628\\
		
		+VSR-Net & \textbf{0.9653} & \textbf{0.8214} & \textbf{0.6627} & \textbf{0.1726} & \textbf{0.1711} & \textbf{0.2509} & \textbf{0.0347}\\
		\hline
	\end{tabular}
  \label{table:table7}
\end{table*}

\begin{figure}[!]
	\centerline{\includegraphics[width=0.8\columnwidth]{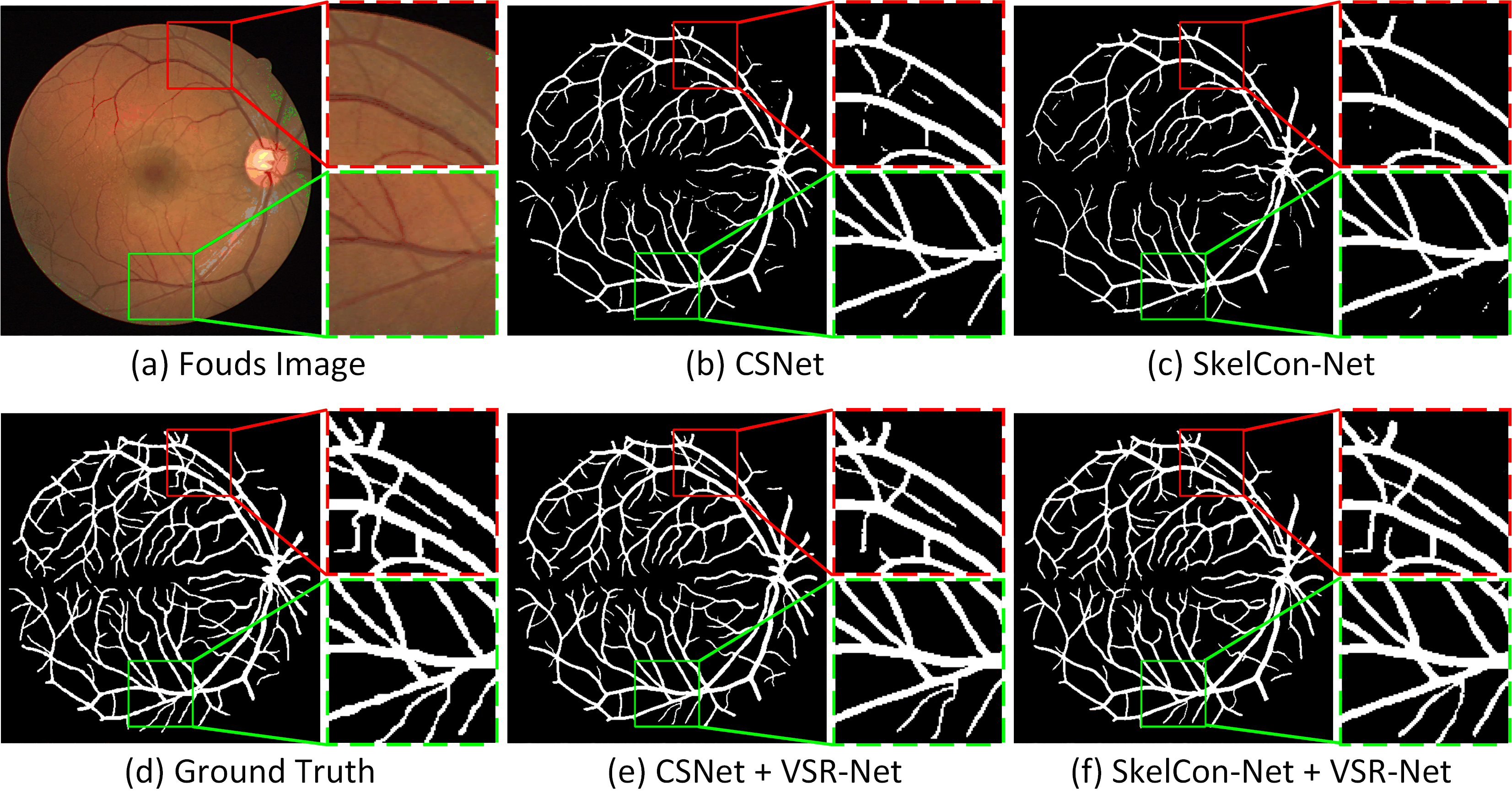}}
	\caption{Visual comparison between vessel-like structure segmentation results by baseline methods and vessel-like structure rehabilitation results by integrating VSR-Net in a representative sample from the DRIVE\cite{DRIVE} test set.}
	\label{fig11}
\end{figure}

\begin{figure}[!]
	\centerline{\includegraphics[width=0.8\columnwidth]{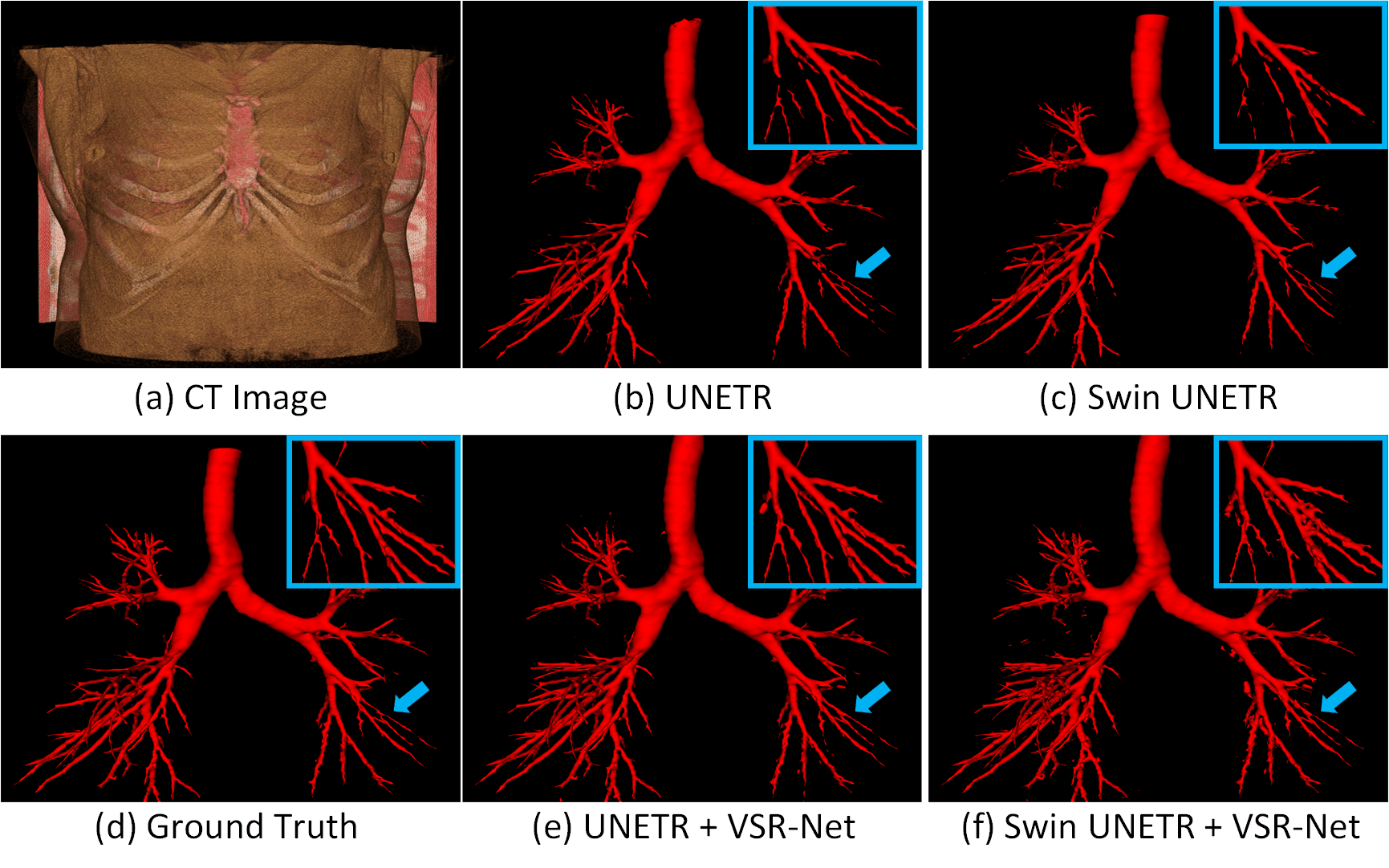}}
	\caption{Visual comparison between vessel-like structure segmentation results by baseline methods and vessel-like structure rehabilitation results by integrating VSR-Net in a representative sample from the ATM\cite{ATM} test set.}
	\label{fig12}
\end{figure}

\subsubsection{Generalization and Reliability Comparison on 2D Vessel-like Structure Datasets}

Table~\ref{table:table6} provides the vessel-like structure segmentation results, vessel-like structure rehabilitation results, and ECE results of original vessel-like structure segmentation baselines and them with our VSR-Net. We conclude as follows: (1) from the segmentation perspective, compared with original baselines, VSR-Net achieves the highest gains of PA, dice score, and Jaccard score are 0.02, 0.0381 and 0.0539, proving our method can further boost vessel-like structure segmentation results. (2) From the vessel-like structure rehabilitation  perspective, VSR-Net significantly reduces the values of VBN, FD, and VT through comparisons to original baselines. Fig. \ref{fig11} shows the visual vessel-like structure rehabilitation comparisons of using VSR-Net to rehabilitate subsection ruptures based on the coarse segmentation results of CS2-Net\cite{CS2Net} and Skelcon-Net\cite{Sklcon}. The rehabilitation process of VSR-Net can also segment some microvessels that were not easily noticeable and not segmented before, improving the integrality of segmentation results. The results demonstrate the superiority of VSR-Net in rehabilitating subsection ruptures by their spatial interconnection relationships via graph clustering. (3) from the confidence calibration perspective, our method also reduces the ECE, validating its reliability. Generally, the results prove the generalization and reliability of VSR-Net in tackling 2D vessel-like structure rehabilitation tasks from three different perspectives.

\subsubsection{Generalization and Reliability  Comparison on 3D Vessel-like Structure Datasets}
We further demonstrate the generalization and reliability of VSR-Net on 3D vessel-like structure rehabilitation tasks, as shown in Table \ref{table:table7}. We observe that VSR-Net significantly outperforms 3D baselines on seven evaluation metrics. For example, VSR-Net obtains the gains of 0.0237, 0.0442 and 0.0464 in PA, dice score, and Jaccard score through comparisons to Swin UNETR~\cite{Sklcon}. Noticeably, our method also reduces VBN, FD, VT and ECE by 0.2555, 0.1228, 0.1861, and 0.0281 accordingly by integrating the VSR-Net into Swin UNETR.
Fig. \ref{fig12} shows the visual vessel-like structure rehabilitation comparisons of using VSR-Net to rehabilitate subsection ruptures based on the coarse vessel-like segmentation results of UNETR\cite{UNETR} and Swin UNETR\cite{SUNETR}. We observe as follows: (1) VSR-Net effectively rehabilitates the rupture subsections based on the coarse segmentation results from UNETR\cite{UNETR} and Swin UNETR\cite{SUNETR},  demonstrating the effectiveness of VSR-Net in 3D vessel-like structure rehabilitation tasks; (2) VSR-Net can robustly rehabilitate rupture subsections from trachea of different thicknesses, which further demonstrates its generalization and reliability. Overall, VSR-Net demonstrated its generalization and reliability in 2D/3D vessel-like structure rehabilitation tasks by exploiting spatial interconnection relationships among subsection ruptures via graph clustering, agreeing with our expectations.

\section{Conclusion}
In this paper, we provide a novel structure rehabilitation perspective to understand and tackle subsection rupture problems in vessel-like structure segmentation tasks by reconsidering it as the vessel-like structure rehabilitation task. Our proposed VSR-Net can effectively rehabilitate subsection ruptures and improve the confidence calibration under this view based on the coarse vessel-like structure segmentation results, by exploring spatial interconnection relationships among subsection ruptures with graph clustering. The extensive experiments on five 2D/3D vessel-like structure datasets manifest the effectiveness of our method. 

To further demonstrate the generalization and reliability of our VSR-Net qualitatively and quantitatively, we conduct sound experiments to prove our method is capable of keeping a promising balance among vessel-like structure segmentation results, vessel-like structure rehabilitation results, and confidence calibration of deep networks, keeping consistent with our motivation. In future work, we plan to extend the proposed method to unsupervised vessel-like structure rehabilitation tasks and provide more theoretical analysis.


\bibliographystyle{IEEEtran}
\bibliography{ref.bib}


 
\vspace{11pt}





\end{document}